\documentclass[AMS,STIX1COL]{WileyNJD-v2}

\articletype{Original article}

\received{1 April 2025}
\revised{XXXX}
\accepted{YYYY}

\usepackage{comment}
\usepackage{cleveref}
\usepackage{tikz}
\usepackage{subcaption}
\usepackage{enumitem}
\newcommand{\gtapprox}{\mathrel{\gtrsim}}

\newcommand{\sM}{\begin{array}{ccccccccc}}
\newcommand{\eM}{\end{array}}

\newcommand{\pd}[2]{\displaystyle\frac{\displaystyle\partial #1}{\displaystyle\partial #2}}

\newcommand{\lb}{\left(}
\newcommand{\rb}{\right)}

\newcommand{\la}{\langle}
\newcommand{\ra}{\rangle}

\newcommand{\sv}{\lb\begin{array}{ccccccccccccccccc}}
\newcommand{\sV}{\begin{bmatrix}}
\newcommand{\eV}{\end{bmatrix}}
\newcommand{\ev}{\end{array}\rb}

\newcommand{\fempty}[1]{{}}

\newcommand{\sty}[1]{{\boldsymbol{#1}}}

\newcommand{\styy}[1]{{\mathbb{#1}}}

\newcommand{\fe}{\sty{ e}}

\newcommand{\fq}{\sty{ q}}

\newcommand{\fs}{\sty{ s}}
\newcommand{\ft}{\sty{ t}}
\newcommand{\fu}{\sty{ u}}

\newcommand{\fx}{\sty{ x}}
\newcommand{\fy}{\sty{ y}}

\newcommand{\fA}{\sty{ A}}
\newcommand{\fB}{\sty{ B}}

\newcommand{\fD}{\sty{ D}}

\newcommand{\fS}{\sty{ S}}

\newcommand{\fX}{\sty{ X}}

\newcommand{\fZ}{\sty{ Z}}

\newcommand{\zero}{\sty{ 0}}
\newcommand{\fzero}{\sty{ 0}}
\newcommand{\ffA}{\styy{ A}}
\newcommand{\ffB}{\styy{ B}}
\newcommand{\ffC}{\styy{ C}}

\newcommand{\ffR}{\styy{ R}}
\newcommand{\ffS}{\styy{ S}}

\newcommand{\ffY}{\styy{ Y}}

\newcommand{\fkappa}{\mbox{\boldmath $\kappa$}}

\newcommand{\fsigma}{\mbox{\boldmath $\sigma$}}

\newcommand{\feps}{\mbox{\boldmath $\varepsilon $}}
\newcommand{\eps}{\varepsilon}

\newcommand{\cA}{{\cal A}}

\newcommand{\cO}{{\cal O}}

\newcommand{\cV}{{\cal V}}

\newcommand{\fbeps}{\ol{\feps}}
\newcommand{\fbsigma}{\ol{\fsigma}}

\newcommand{\WT}[1]{\widetilde{#1}}
\newcommand{\WH}[1]{\widehat{#1}}
\newcommand{\ol}[1]{\overline{#1}}

\renewcommand{\ul}[1]{\underline{#1}} 
\newcommand{\ulWH}[1]{\ul{\widehat{#1}}}
\newcommand{\ulWT}[1]{\ul{\widetilde{#1}}}

\newcommand{\ull}[1]{\ul{\ul{#1}}}
\newcommand{\ullWH}[1]{\ull{\widehat{#1}}}
\newcommand{\ullWT}[1]{\ull{\widetilde{#1}}}
\newcommand{\ullWTH}[1]{\WH{\ullWT{#1}}}
\newcommand{\ulWTH}[1]{\WH{\ulWT{#1}}}

\newcommand\python[1]{\colorbox{lightgray!30}{\texttt{#1}}}

\newcommand{\Voigt}[1]{{#1}_\mathrm{V}}
\newcommand{\Reuss}[1]{{#1}_\mathrm{R}}
\newcommand{\Cvoigt}{\Voigt{\ol{\ffC}}}
\newcommand{\Creuss}{\Reuss{\ol{\ffC}}}

\newcommand{\Sreuss}{\Reuss{\ol{\ffS}}}

\newcommand{\ffbC}{\ol{\ffC}}
\newcommand{\ffbS}{\ol{\ffS}}
\newcommand{\VRNet}{{Voigt-Reuss net}}

\definecolor{uniSlightblue}{HTML}{00BEFF}
\definecolor{uniSblue}{HTML}{004191}
\definecolor{uniSlblue}{HTML}{00BEFF}
\colorlet{uniSlblue40}{uniSlblue!40!white}
\definecolor{uniSgray}{RGB}{62, 68, 76}
\colorlet{uniSgray20}{uniSgray!20!white}

\begin{document}

\title{Spectral Normalization and \VRNet{}: A universal approach to microstructure-property forecasting with physical guarantees}

\author[1]{Sanath Keshav}

\author[1]{Julius Herb}

\author[1]{Felix Fritzen*}

\authormark{KESHAV \textsc{et al}}

\address[1]{\orgdiv{SC Simtech, Data Analytics in Engineering}, \orgname{University of Stuttgart}, \orgaddress{\state{Universitätsstr.32, 70569 Stuttgart}, \country{Germany}}}

\corres{* Felix Fritzen, Institute of Applied Mechanics (MIB), Universitätsstraße 32, 70569 Stuttgart, Germany. \email{fritzen@mib.uni-stuttgart.de}}

\abstract[Summary]{Heterogeneous materials are crucial to producing lightweight components, functional components, and structures composed of them. A crucial step in the design process is the rapid evaluation of their effective mechanical, thermal, or, in general, constitutive properties. The established procedure is to use forward models that accept microstructure geometry and local constitutive properties as inputs. The classical simulation-based approach, which uses, e.g., finite elements and FFT-based solvers, can require substantial computational resources. At the same time, simulation-based models struggle to provide gradients with respect to the microstructure and the constitutive parameters. Such gradients are, however, of paramount importance for microstructure design and for inverting the microstructure-property mapping. Machine learning surrogates can excel in these situations. However, they can lead to unphysical predictions that violate essential bounds on the constitutive response, such as the upper (Voigt-like) or the lower (Reuss-like) bound in linear elasticity. Therefore, we propose a novel spectral normalization scheme that \textit{a priori} enforces these bounds. The approach is fully agnostic with respect to the chosen microstructural features and the utilized surrogate model: It can be linked to neural networks, kernel methods, or combined schemes. All of these will automatically and strictly predict outputs that obey the upper and lower bounds by construction. The technique can be used for any constitutive tensor that is symmetric and where upper and lower bounds (in the Löwner sense) exist, i.e., for permeability, thermal conductivity, linear elasticity, and many more. We demonstrate the use of spectral normalization in the \VRNet{} using a simple neural network. Numerical examples on truly extensive datasets illustrate the improved accuracy, robustness, and independence of the type of input features in comparison to much-used neural networks.}


\JELinfo{classification}

\MSC{Code numbers}

\jnlcitation{\cname{%
\author{S. Keshav},
\author{J. Herb}, and
\author{F. Fritzen}} (\cyear{2025}),
\ctitle{Spectral Normalization and \VRNet{}: A universal approach to microstructure-property forecasting with physical guarantees}, \cjournal{XXX}, \cvol{XXX}.}

\maketitle

\section{Introduction}

The aim of multiscale materials modeling is to uncover the hidden relation between the microstructure and the effective behavior of the heterogeneous materials at a larger scale. The term microstructure refers to the presence of features that are much smaller than the length scale of the engineering application of the material. Microstructural features comprise inclusions, pores, or crystallographic grains. They can strongly influence effective properties such as thermal conductivity, elasticity, permeability, and other physical properties. Homogenization techniques bridge scales by forecasting effective constitutive properties based on assumptions on the microstructure. For instance, computational schemes make use of a representative volume element (RVE) of the microstructure. The size of the RVE must be chosen carefully \cite{kanit2003,ostoja2006}. The computations on the RVE level can be based on simplifying mean-field theories~\cite{chaboche2005} and full-field strategies based on, e.g., finite element \citep[FE; e.g.][]{wriggers2008} and Fast Fourier Transform (FFT) based methods \citep{moulinec1998,Leuschner2017,Schneider2021} to solve the microscale boundary value problem and extract effective behavior. These simulation-based methods have become the de-facto standard in computational micromechanics, offering high-fidelity predictions given a detailed microstructure description \cite{zohdi2005}. However, they come with well-known limitations: (i) the computational cost is prohibitively high, especially for nonlinear or high-resolution 3D analyses, and (ii) they behave as black-box forward solvers lacking analytical gradients with respect to microstructural features or constituent material parameters. The latter is particularly restrictive for design tasks since one cannot easily explore the microstructural composition space using direct simulations. As a result, optimizing microstructures for target properties often requires brute-force searches or heuristic optimization on top of many forward simulations, which is intractable in large design spaces.

Advancing computational methods to allow the inversion of the microstructure-property relation has been achieved in recent years through machine learning (ML) surrogates, which have emerged as a promising alternative to direct numerical simulations to accelerate and generalize multiscale modeling. Instead of performing a full physical simulation for each new microstructure, ML models can be trained on data (from experiments, high-fidelity simulations, or combinations thereof) to learn a mapping from microstructural descriptors to the effective properties. Once trained, such a surrogate can predict properties really fast, even in real-time, enabling fast exploration of the microstructure design space. Moreover, ML models (especially neural networks) are differentiable with respect to the inputs (often called input features) by construction by utilizing the power of automatic differentiation \cite{baydin2018automatic}. This enables the computation of sensitivities of the effective response with respect to the material composition either through a reduced set of microstructural descriptors or even with respect to the image of a material. The availability of such sensitivities is crucial for the efficient optimization of microstructured solids \citep[e.g.,][]{Peng2024_inverse,Rassloff2025}.

With respect to machine learning for the microstructure-informed forward prediction, Deep Material Networks (DMNs) were introduced as a physics-guided neural architecture for multiscale topology learning, demonstrating how a hierarchical neural network can learn the RVE behavior while respecting some mechanistic principles \cite{Liu2019}. Subsequent works refined and applied this concept to complex composites, showing significant speed-ups in two-scale simulations \cite{Gajek2020,Wu2025}. More general methods that do not directly incorporate mechanistic principles have been proposed by researchers using different types of neural networks (feed-forward neural nets, convolutional neural nets), Gaussian process regressions~\cite{Liu2020}, and other ML techniques \citep[e.g., kernel-based methods,][]{scholkopf2002learning} to predict effective constitutive responses from microstructure characteristics. Comprehensive reviews of these efforts highlight the growing interest and success of ML in constitutive modeling and homogenization \citep[e.g.,][]{Liu2021, Peng2024}. Indeed, by replacing expensive FE/FFT queries with a learned model, one can integrate multiscale evaluations directly into optimization loops or perform on-the-fly material screening, which was previously impractical.

Despite their advantages, purely data-driven surrogates can suffer from a lack of physical constraints, potentially yielding unphysical or inconsistent predictions. Standard ML models have no built-in guarantee of obeying fundamental physics laws or material principles. In the context of constitutive modeling, this issue manifests in several ways. For instance, a naive neural network predictor for elastic stiffness might violate the symmetry of the stiffness tensor or even predict a matrix that is not positive-definite, thus breaking thermodynamic requirements (e.g., a stiffness that allows negative energy or non-unique responses). Similarly, ML models might extrapolate beyond theoretical limits. For instance, they might predict the stiffness of a composite to be stiffer than the established and strict bounds~\cite{Voigt1889,Reuss1929,Wiener1912}.

The enforcement of physical properties can be approximated by penalizing violations of physical properties \citep[physics informed neural networks;][]{Cuomo2022}, much like material overclosure can be prevented by a high contact stiffness. This weak enforcement, however, can struggle with the selection of the penalty parameter and can seldom reduce the amount of needed input data.
These concerns have motivated a new wave of research on physics-aware ML models \cite{Karniadakis2021,Farhat2022,Linden2023}. The central idea in such models is the incorporation of physical prior knowledge into the learning process. Thereby, surrogate models can be forced to respect known laws by design rather than only through data. A positive side-effect is the reduction of the amount of needed data, a much-desired property in view of the expensive data generation through simulation.

One branch of physics-aware models enforces thermodynamic consistency and material stability expressed through qualitative mathematical properties of the constitutive model. For example, in data-driven hyperelasticity, networks have been designed to represent an underlying strain energy potential that is polyconvex or convex in strains, using architectures like input-convex neural networks, thereby guaranteeing realistic stress–strain behavior \cite{Farhat2022,Linden2023}. Recently, it has been noted that polyconvexity enforced through input convex neural networks \cite{icnn} might over-constrain the strain energy function, leading to inaccurate models \cite{Klein2025}. This can be overcome by relaxing the constraints; see also \cite{Kalina2024}.

Physics-aware surrogates can also be achieved on a lower level, e.g., by enforcing the symmetry and positive definiteness of constitutive tensors. Xu et al.~\cite{Xu2021} introduced a Cholesky-factored neural network architecture that always outputs a symmetric positive-definite (SPD) stiffness matrix by predicting a factor (like a square root of the tensor) and thus ensuring material stability by construction. Likewise, Huang et al.~\cite{huang2017riemannian} developed a Riemannian deep network on the manifold of SPD matrices to better learn mappings involving covariance matrices, illustrating how respecting geometric structure in the model can improve performance. In multiscale modeling, material symmetry groups \cite{Fernandez2021} can be built into the model, and networks like the DMN inherently respect certain mixture rules by construction \cite{Liu2019, Gajek2020}. These efforts collectively show that introducing physics knowledge- be it through architecture, training constraints, data augmentation, or feature design- can greatly improve the robustness and credibility of ML-based constitutive models. Still, most existing approaches focus on \textit{qualitative} physical requirements (symmetry, positivity, invariance, etc.), whereas \textit{quantitative} theoretical limits provided by analytical homogenization theory have received comparatively little attention in surrogate modeling.

In particular, a distinguishing contribution of this work is the strict prior enforcement of upper (Voigt-like) and lower (Reuss-like) bounds within a machine-learning model for effective property prediction. The Voigt and Reuss bounds are classical analytical estimates that bound the effective moduli of composite material, assuming uniform strain and uniform stress conditions, respectively, in the RVE \citep[see, e.g.,][]{nemat2013micromechanics,zohdi2005}. They provide rigorous limits: no real heterogeneous microstructure can have an effective stiffness above the Voigt bound or below the Reuss bound (and similarly for conductivity, permeability, etc.). However, generic ML models have no awareness of these bounds and could easily predict constitutive tensors violating them. To our knowledge, this work is the first to explicitly embed Voigt- \textit{and} Reuss-like constraints simultaneously into a learning model for constitutive properties. By doing so, we ensure that the surrogate cannot yield a physically impossible prediction, no matter the input. We achieve this through a novel spectral normalization technique. This procedure guarantees that (a) the output tensor is always symmetric, and (b) the effective property tensor respects both upper and lower bounds in the Löwner sense. The framework is quite general: It can be wrapped around different regression models or all sorts of neural networks without requiring a specific architecture. For example, one could use standard ("deep") neural networks, Gaussian processes, kernel regressors, or a hybrid physics–ML model as the base predictor. This modular nature aligns with the philosophy of physics-aware ML in that it augments any learning model with a physics-based adjustment of the output manifold. In summary, our approach merges the data-fitting capacity of modern ML with the hard constraints of homogenization theory, yielding a fast, differentiable, and physics-aware surrogate model.

In \Cref{sec:spectral:normalization} the spectral normalization is described starting from a mechanical motivation, which is then restricted to 1D to sketch the underlying ideas of the general approach described in \Cref{subsec:spectral:normalization:general}. We then show how the spectral normalization can be implemented into a general data-driven model, in our case into \texttt{pytorch}~\cite{Ansel_PyTorch_2_Faster_2024}, using a simple feed-forward neural network architecture. \Cref{sec:results} then presents extensive 2D and 3D datasets for thermal homogenization of binary microstructures. It is demonstrated how the proposed method can significantly leverage the accuracy of the data-driven model independent of the number and type of input features.

\subsection{Notation}
The spatial average of a quantity over a domain $\cA$ with measure $A = \lvert \cA \rvert$ is defined as 
\begin{equation}
	\langle\cdot\rangle_{\cA}=\dfrac{1}{A} \int_{\cA}(\cdot) \; \mathrm{d} A \, .
\end{equation}
In the sequel, boldface lowercase letters denote vectors, boldface upper case letters denote 2-tensors, and blackboard bold uppercase letters (e.g., $\ffC$) denote 4-tensors. We use single and double contractions $\cdot, :$. With respect to an orthonormal basis, they are defined as
\begin{align*}
\fA \cdot \fB &= A_{\cdots i} B_{i \cdots}, & \fA : \fB &= A_{\cdots ij} B_{ij \cdots},
\end{align*}
using Einstein summation over repeated indices. The tensor product is denoted by $\otimes$. Using the above notation in conjunction with the operator
\begin{align*}
    \nabla &= \sV \partial_x \\ \partial_y \\ \partial_z \eV \, ,
\end{align*}
the divergence and the (right) gradient of a general tensor read
\begin{align*}
    \fA \cdot \nabla &= A_{\cdots i, i}, &
    \fA \otimes \nabla &= A_{\cdots i, j} \, .
\end{align*}
For convenience, a six-dimensional orthonormal basis of the space of symmetric second-order tensors in a Mandel-like notation is proposed:
\begin{align*}
    \fB_1 &= \fe_1 \otimes \fe_1, &
    \fB_2 &= \fe_2 \otimes \fe_2, &
    \fB_3 &= \fe_3 \otimes \fe_3 \\
    \fB_4 &= \frac{\sqrt{2}}{2} \lb \fe_1 \otimes \fe_2 + \fe_2 \otimes \fe_1 \rb, &
    \fB_5 &= \frac{\sqrt{2}}{2} \lb \fe_1 \otimes \fe_3 + \fe_3 \otimes \fe_1 \rb, &
    \fB_6 &= \frac{\sqrt{2}}{2} \lb \fe_2 \otimes \fe_3 + \fe_3 \otimes \fe_2 \rb \, .
\end{align*}
Then any symmetric second-order tensor $\fs$ and any fourth-order tensor~$\ffY$ with left and right sub-symmetry can be expressed by a vector $\ul{s}$ and a matrix $\ull{Y}$ via
\begin{align*}
    s_i &= \fS : \fB_i, &
    Y_{ij} &= \fB_i : \ffY : \fB_j \, .
\end{align*}
The following identities hold for symmetric tensors of order 2 ($\fs, \ft$) and 4 ($\ffC, \ffS$) and their respective vector representations:
\begin{align*}
    \fs \cdot \ft & \leftrightarrow \ul{s}^\mathsf{T} \ul{t} = \ul{s} \cdot \ul{t} = \ul{t}^\mathsf{T} \ul{s}, &
    \ffC : \fs & \leftrightarrow \ull{C} \, \ul{s}, &
    \ffC^{-1} : \ft & \leftrightarrow \ull{C}^{-1} \, \ul{t} \, .
\end{align*}

\section{Spectral normalization}
\label{sec:spectral:normalization}

\subsection{Prerequisite: Symmetry and positivity of constitutive tensors}
Many constitutive tensors have notable mathematical properties inherited from basic physical principles: In many situations, a constitutive tensor $\ffY$ mapping an input $\fX$ to an output $\fZ$ is both symmetric and positive definite:
\begin{align}
    \fZ \cdot \ffY \cdot \fX &= \fX \cdot \ffY \cdot \fZ \, , &
    \fX \cdot \ffY \cdot \fX & > 0 \, . \label{eq:prerequisite}
\end{align}
Here, the operator $\cdot$ denotes the full contraction (diverging slightly from the notation owing to the unconstrained order of the tensors).
Examples comprise:
\begin{itemize}
    \item The thermal conductivity tensor $\fkappa$ is symmetric by Onsager's reciprocity theorem and positive definite to guarantee non-negative thermal dissipation.
    \item The elasticity tensor $\ffC$ is symmetric by the Schwarz theorem, as it is a second gradient of a strain energy density. In addition, the energy of deformed elastic materials is greater than that of undeformed materials, inducing the positivity of $\ffC$.
\end{itemize}
Likewise, diffusivity $\fD$, permeability $\fA$, and related quantities meet the prerequisite~\cref{eq:prerequisite}.
\subsection{Motivation: Strict upper and lower bounds in linear elasticity}
In a linear elastic medium, the symmetric gradient of the displacement~$\fu$ defines the infinitesimal strain tensor~$\feps(\fx)=\text{sym}(\fu(\fx) \otimes \nabla)$. The strain~$\feps$ is related to the stress~$\fsigma(\fx)$ through the elasticity tensor~$\ffC(\fx)$ via
\begin{align}
    \fsigma(\fx) &= \ffC(\fx) : \feps(\fx)  = \pd{W(\feps, \fx)}{\feps} \, ,
\end{align}
where $W(\feps, \fx)$ is a strain energy density of the form
\begin{align}
    W(\feps, \fx) &= \frac{1}{2}\; \feps(\fx) : \ffC(\fx) : \feps(\fx) \, .
\end{align}
When dealing with microstructured materials defined by a representative volume element~$\Omega$, the displacement field~$\fu$ is constrained by the applied kinematic loading, i.e., by the applied macroscopic strain $\fbeps$:
\begin{align}
    \fu & \in \cV_\square(\fbeps) = \{ \fu = \fbeps \cdot \fx + \WT{\fu}, \ \la \WT{\fu} \ra = \zero, \ \la \WT{\fu} \otimes \nabla \ra = \zero \} \, . \label{eq:u:admissible}
\end{align}
The term $\WT{\fu}$ represents displacement fluctuations, i.e., deviations from a homogeneous deformation of the material. The total strain energy of the deformed microstructure given an admissible displacement~$\fu\in\cV_\square$ is
\begin{align}
    \ol{\Pi}( \fu, \fbeps ) &= \frac{1}{\vert\Omega\vert} \intop_\Omega W(\feps(\fu), \fx ) {\rm d} \Omega , &
    \feps(\fu) &= \text{sym} ( \fu \otimes \nabla ) = \fbeps + \text{sym} ( \WT{\fu} \otimes \nabla ), &
    \fu & \in \cV_\square(\fbeps) \, .
\end{align}
Resorting to the virtual work principle, the minimizer $\fu_*\in\cV_\square(\fbeps)$ of $\ol{\Pi}$ yields the solution to the elasticity problem for given~$\fbeps$. It defines the total strain energy as a function of just~$\fbeps$:
\begin{align}
    \fu_\ast &= \underset{\fu\in\cV_\square(\fbeps)}{\text{arg min}} \ol{\Pi}(\fu, \fbeps) \, , &
    \ol{W}(\fbeps) &= \ol{\Pi}(\fu_\ast, \fbeps) \, .
\end{align}
Consequently, \textit{any} field $\fu \in \cV_\square(\fbeps)$ satisfies
\begin{align}
    \ol{\Pi}(\fu, \fbeps) & \geq \ol{\Pi} (\fu_\ast, \fbeps) = \ol{W}(\fbeps) \, .
\end{align}
Notably, we can write
\begin{align}
    \fu &= \fbeps \cdot \fx + \text{sym}(\WT{\fu} \otimes \nabla) \in \cV_\square(\fzero) & \Leftrightarrow \ \WT{\fu} & \in \cV_\square(\fzero) \, .
\end{align}
Since $\WT{\fu}=\fzero \in \cV_\square(\fzero)$ holds independent of $\fbeps$, a trivial upper bound on $\ol{W}$ is
\begin{align}
    \Voigt{\ffbC} &= \frac{1}{\vert\Omega\vert} \intop_\Omega \ffC(\fx) {\rm d}\Omega \, , \label{eq:Cvoigt} \\
    \ol{\Pi} ( \fzero, \fbeps ) &= \frac{1}{2\vert\Omega\vert} \intop_\Omega \fbeps : \ffC(\fx) : \fbeps \, {\rm d} \Omega
    = \frac{1}{2} \; \fbeps : \Voigt{\ffbC} : \fbeps \, . \label{eq:el:upper}
\end{align}
By the linearity of the problem, the effective strain energy density~$\ol{W}(\fbeps)$ inherits the structure of the local strain energy~$W(\feps)$:
\begin{align}
    \ol{W}(\fbeps) &= \frac{1}{2} \; \fbeps : \ffbC : \fbeps \, ,\label{eq:Wbar}
\end{align}
where $\ffbC$ is the sought-after unknown effective elasticity tensor.
Combining \cref{eq:el:upper} and \cref{eq:Wbar} leads to
\begin{align}
 \frac{1}{2} \;\fbeps : \ffbC : \fbeps  =  \ol{W}(\fbeps)  &\leq \ol{\Pi}(\fzero, \fbeps) = \frac{1}{2} \; \fbeps : \Voigt{\ffbC} : \fbeps \, .
\end{align}
This inequality holds independent of the applied strain $\fbeps$, resulting in the Löwner order
\begin{align}
    \ffbC & \preceq \Voigt{\ffbC} \ \Leftrightarrow \ \fbeps : \lb \Voigt{\ffbC} - \ffbC \rb : \fbeps  \geq 0 \quad \forall \;\fbeps \in Sym(\ffR^{3\times 3}) \, . \label{eq:Voigt:order}
\end{align}
The tensor $\Cvoigt$ is called the Voigt estimate. It denotes an upper bound to the effective stiffness tensor~$\ffbC$ with no assumptions imposed regarding the geometry of the microstructured material; see, e.g., \cite{nemat2013micromechanics} for further details.

Instead of working with the strain energy $W(\feps, \fx)$, a stress-centered approach based on the complementary energy 
\begin{align}
    W^\ast ( \fsigma, \fx) &= \frac{1}{2}\; \fsigma(\fx) : \ffS(\fx) : \fsigma(\fx)
\end{align}
using the compliance tensor $\ffS(\fx)=\ffC^{-1}(\fx)$ can be chosen. Here, $\fsigma(\fx)$ is a stress field that is statically admissible for a given macroscopic stress~$\fbsigma$, i.e., which satisfies\footnote{On purpose, we do not dive into the definition of the function space from which $\fsigma$ stems and rather focus on the relevant content. In short, $\fsigma$ must be piecewise continuously differentiable. Furthermore, the traction jump on points with discontinuous derivatives must be zero.}
\begin{align}
    \fsigma(\fx) \in \cV^\ast_\square ( \fbsigma) = \left\lbrace \fsigma(\fx)\text{ s. th. }\fsigma(\fx) \cdot \nabla = \fzero, \ \left\la \fsigma(\fx) \right\ra = \fbsigma \right\rbrace \, . \label{eq:sigma:admissible}
\end{align}
The complementary energy is defined for $\fsigma\in\cV^\ast_\square(\fbsigma)$ via
\begin{align}
    \ol{\Pi}^\ast(\fsigma, \fbsigma) &= \frac{1}{\vert\Omega\vert} \intop_\Omega W^\ast(\fsigma, \fx) \;{\rm d} \Omega = \frac{1}{2\vert\Omega\vert} \intop_\Omega \fsigma(\fx) : \ffS (\fx) : \fsigma(\fx) \; {\rm d}\Omega \, .
\end{align}
The minimizer~$\fsigma_\ast\in\cV^\ast_\square$ of $\ol{\Pi}^\ast(\fsigma, \fbsigma)$ defines the effective complementary energy~$\ol{W}(\fbsigma)$
\begin{align}
    \fsigma_* &= \underset{\displaystyle \fsigma\in\cV^\ast_\square}{\text{arg max}} \  \ol{\Pi}^\ast(\fsigma, \fbsigma) , &
    \ol{W}^\ast(\fbsigma) &= \left\la W^\ast(\fsigma^\ast, \fx) \right\ra \leq \ol{\Pi}^\ast(\fsigma, \fbsigma) \quad \forall \fsigma\in\cV^\ast_\square(\fbsigma) \, .
\end{align}
Using the representation of $\ol{W}^\ast$ in terms of the effective compliance $\ffbS={\ffbC}^{-1}$ gives
\begin{align}
    \ol{W}^\ast(\fbsigma) &= \frac{1}{2}\; \fbsigma : \ffbS : \fbsigma \, .
\end{align}
Obviously, $\ol{W}^\ast(\fbsigma)$ is smaller or equal than $\ol{\Pi}^\ast (\fsigma, \fbsigma)$ for \textit{any} $\fsigma\in\cV^\ast_\square(\fbsigma)$. Similar to the Voigt bound one can choose $\fsigma(\fx)=\fbsigma \in \cV^\ast_\square(\fbsigma)$ to gain
\begin{align}
    \Creuss^{-1} = \Sreuss &= \frac{1}{\vert\Omega\vert} \intop_\Omega \ffS(\fx) \, {\rm d} \Omega , & \label{eq:Creuss}
    \fbsigma : \ffbS : \fbsigma & \leq \fbsigma : \Reuss{\ffbS} : \fbsigma \, .
\end{align}
Due to the arbitrariness of $\fbsigma$, the following Löwner order is implied:
\begin{align}
\ffbS &= \ffbC^{-1} \preceq \Reuss{\ffbS} =  \Reuss{\ffbC}^{-1} &
\Leftrightarrow \ \ffbC & = \ffbS^{-1} \succeq \Reuss{\ffbS}^{-1} = \Reuss{\ffbC} \, . \label{eq:Reuss:order}
\end{align}
Combining \cref{eq:Voigt:order,eq:Reuss:order} the effective stiffness is bounded via
\begin{align}
    \Creuss \preceq \ol{\ffC} \preceq \Cvoigt.
\end{align}
Here $\preceq$ is referring to the L\"owner ordering, i.e.:
\begin{align}
    \ffA & \preceq \ffB \leftrightarrow \ \forall \fy: \ \fy : \ffA : \fy \leq \fy:\ffB : \fy \, .
\end{align}

\begin{remark}
For piece-wise constant stiffness $\ffC_i$ in the $i^\text{th}$ material with volume fraction $c_i $ where $0 \leq c_i \leq 1$ and $\sum_{i=1}^n c_i = 1$, the Voigt estimate~\cref{eq:Cvoigt} simplifies to
\begin{align}
    \ol{\ffC}_\mathrm{V} & = \sum_{i=1}^n c_i \ffC_i \, .
\end{align}
In the same way, the Reuss estimate~\cref{eq:Creuss} can be expressed via
\begin{align}
    \Sreuss &=  \sum_{i=1}^n c_i \ffC_i^{-1}, &
    \Creuss &= \Sreuss^{-1} = \lb \sum_{i=1}^n c_i \ffC_i^{-1} \rb^{-1}\, .
\end{align}
\end{remark}

\begin{remark}
The Voigt and Reuss bounds are first-order bounds, i.e., they depend only on the volume fraction but not on the morphology and topology of phases. An even rougher estimate can be obtained via the 0$^\text{th}$ order bound:
\begin{align}
    \ffC_i & \preceq \ol{\ffC} \preceq \ffC_i \qquad \forall i \in \{ 1, \dots, n \}.
\end{align}
The physical interpretation of the bounds of 0$^\text{th}$ order is that the stiffness will be smaller than that of the stiffest component and higher than that of the most compliant one.
\end{remark}

\begin{remark}
Although originally formulated in the framework of linear elasticity, the Voigt and Reuss bounds generalize to many physical problems that are described by elliptic partial differential equations. In case of (stationary) heat conduction, the heat flux $\fq(\fx)$ is related to the temperature $\theta(\fx)$ via the symmetric and positive definite conductivity tensor $\fkappa$ through
\begin{align}
    \fq(\fx) &= - \fkappa(\fx) \cdot ( \theta(\fx) \otimes \nabla) \, .
\end{align}
The bounds on $\ol{\fkappa}$ are then given via
\begin{align}
    \Voigt{\ol{\fkappa}} &= \frac{1}{\vert\Omega\vert} \intop_\Omega \fkappa(\fx) \, \mathrm{d} \Omega\, ,  &
    \Reuss{\ol{\fkappa}} &= \lb \frac{1}{\vert\Omega\vert} \intop_\Omega \fkappa^{-1}(\fx) \, \mathrm{d} \Omega \rb^{-1}\, . \label{eq:VoigtReuss:kappa}
\end{align}
\end{remark}

\subsection{Spectral normalization}

\subsubsection{One-dimensional motivation}
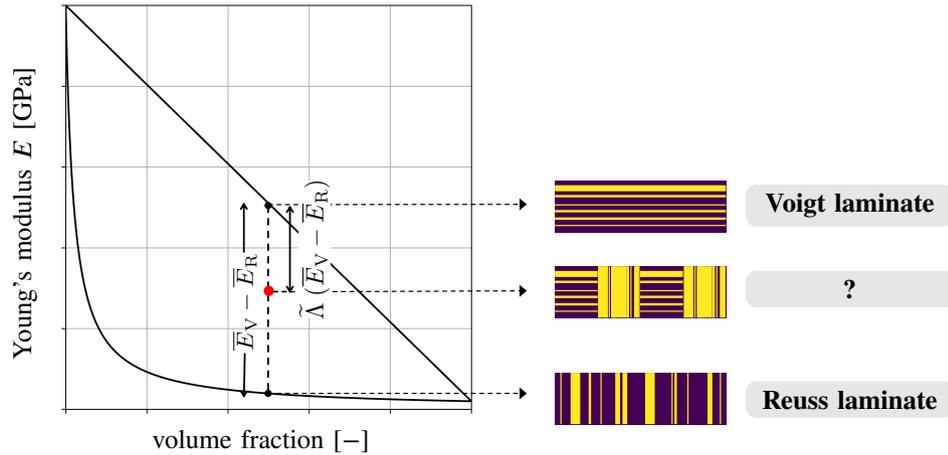
\begin{figure}[ht!]
    \centering
    \includegraphics[page=5,scale=0.45,trim={5.5cm 1.75cm 2.5cm 1cm},clip]{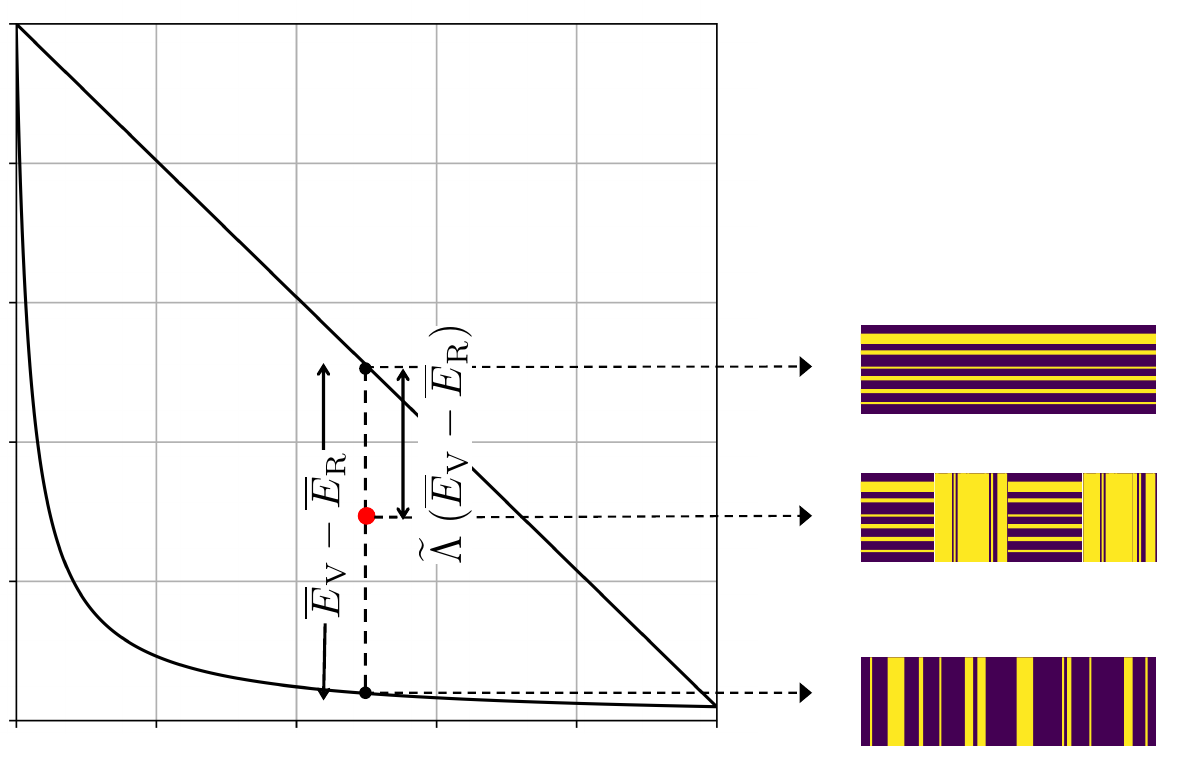}
    \begin{tikzpicture}[overlay, remember picture]
        \node[text width=3.5cm, align=center, rotate=0, rounded corners] at (-6.5,0.0) {volume fraction $[-]$};
        \node[text width=4.0cm, align=center, rotate=90, rounded corners] at (-9.6,3.0) {Young's modulus $E$ [GPa]};
        \node[fill=gray!20, text width=2.5cm, align=center, rotate=0, rounded corners] at (1.25,3.1) {\textbf{Voigt laminate}};
        \node[fill=gray!20, text width=2.5cm, align=center, rotate=0, rounded corners] at (1.25,2.0) {\textbf{?}};
        \node[fill=gray!20, text width=2.5cm, align=center, rotate=0, rounded corners] at (1.25,0.55) {\textbf{Reuss laminate}};
    \end{tikzpicture}
    \vspace{0.5cm}
    \caption{Schematic representation of the one-dimensional laminate structure and its effective Young's moduli.}
    \label{fig:1d_laminate}
\end{figure}

Suppose we are eyeing a heterogeneous material composed of two phases with unknown spatial arrangement; see \Cref{fig:1d_laminate}. The volume fractions are $c_1, c_2 \in (0, 1)$, $c_1 + c_2 = 1$ and Young's modulus is $E_1$ and $E_2$ in either phase, respectively. We now consider the two extremal cases in which the materials could be aligned in a parallel arrangement ($\bullet_\mathrm{par}$) or a serial arrangement ($\bullet_\mathrm{ser}$). Via a parallel arrangement, i.e., pulling in the direction of the lamella, the stress~$\ol{\sigma}$ given a uniaxial strain $\ol{\varepsilon}$ add up according to:
\begin{align}
\ol{\sigma}_\mathrm{par} &= \lb c_1 E_1 + c_2 E_2 \rb \ol{\varepsilon} = \Voigt{\ol{E}} \ol{\varepsilon} \, .
\end{align}
The modulus $\Voigt{\ol{E}}$ denotes an upper bound on the effective modulus $\ol{E}$.

Next, we consider the other extreme, namely a serial arrangement of the materials in the direction of loading. When applying a strain in the perpendicular direction, the stresses in both materials must match in order to be statically admissible:
\begin{align}
    \sigma_1 &= E_1 \varepsilon_1 = E_2 \varepsilon_2 = \sigma_2 = \ol{\sigma}_\mathrm{ser}\, .
\end{align}
The strains are constrained by
\begin{align}
    c_1 \varepsilon_1 + c_2 \varepsilon_2 &= \ol{\varepsilon} \, .
\end{align}
Combining these two relations, we gain:
\begin{align}
    E_1 \varepsilon_1 = \frac{1}{c_2} E_2 (\ol{\varepsilon} - c_1 \varepsilon_1 ) &
    \Rightarrow \ \varepsilon_1 
    = \frac{E_2}{c_2 E_1 + c_1 E_2} \ol{\varepsilon} \,.
\end{align}
From this, the effective stress gets
\begin{align}
    \ol{\sigma}_\mathrm{ser} &= E_1 \varepsilon_1 = E_2 \varepsilon_2 = \ol{E} \; \ol{\varepsilon} = \frac{E_1 E_2}{c_2 E_1 + c_1 E_2}\; \ol{\varepsilon} = \frac{1}{ E_1/c_1 + E_2 / c_2} \;\ol{\varepsilon} = \Reuss{\ol{E}} \ol{\varepsilon}  \, .
\end{align}
By mixing parallel and sequential arrangements, an intermediate modulus $\ol{E}$ in between the lower Reuss bound $\Reuss{\ol{E}}$ and the upper Voigt bound $\Voigt{\ol{E}}$ is found. Most importantly, they denote strict bounds on the response. In the interest of a non-dimensional description of the actual stiffness $\ol{E}$, we introduce a new parameter $\WT{\Lambda}\in [0, 1]$, such that
\begin{align}
    \ol{E} & = \Voigt{\ol{E}} - \WT{\Lambda}\; (\Voigt{\ol{E}} - \Reuss{\ol{E}}),
\end{align}
where the parameter $\WT{\Lambda}$ depends on the actual arrangement. We propose to exploit the de-dimensionalized parameterization in a machine-learning context as it provides several advantages:
\begin{itemize}
    \item The parameter is independent of the material parameters (\textit{here:} $E_1$, $E_2$).
    \item The parameter is devoid of a physical dimension.
    \item The range is well-defined, and the (absolute) errors in $\WT{\Lambda}$ correspond to relative errors with respect to the range of admissible material responses.
    \item The enforcement of $\WT{\Lambda}\in[0, 1]$ is straightforward. Thereby, it can be guaranteed that physically sound predictions are made \textit{a priori}.
\end{itemize}

\subsubsection{Generalization to arbitrary positive definite tensors} 
\label{subsec:spectral:normalization:general}
In order to generalize the scalar parameter $\WT{\Lambda}$ for symmetric positive definite operators as labeled in \cref{eq:prerequisite}, we start by introducing some technical simplifications:
\begin{itemize}
    \item The tensors $\fX, \fZ$ in \cref{eq:prerequisite} can be expressed in an orthonormal basis by coefficient vectors $\ul{x}, \ul{z} \in \ffR^m$.
    \item Likewise, the tensor $\ffY$ can be expressed by a matrix $\ull{Y} \in \ffR^{m\times m}$.
    \item The symmetry and positive definiteness of $\ffY$ induce $\ull{Y}$ to be symmetric and positive definite, too.
    \item Existence of an upper and lower bound $\Voigt{\ull{Y}}$ and $\Reuss{\ull{Y}}$, respectively.
\end{itemize}
Under the given assumptions and the above notation, the Löwner ordering implies
\begin{align}
    \Reuss{\ull{Y}} & \leq \ull{Y} \leq \Voigt{\ull{Y}} \ \Rightarrow\ 
    0 \leq \Voigt{\ull{Y}} - \ull{Y} \leq \Voigt{\ull{Y}} - \Reuss{\ull{Y}} \, . \label{eq:preparation}
\end{align}
The latter term expresses the difference between the upper and lower bound. By definition, it is symmetric and positive semi-definite. Therefore, a diagonalization can be applied:
\begin{align}
    \ull{Q}_0 \; \ull{\Lambda}_0 \; \ull{Q}_0^\mathsf{T} &=  \Voigt{\ull{Y}} - \Reuss{\ull{Y}}  \, .
\end{align}
In most situations, $\ull{\Lambda}_0$ is positive definite. In case it is not, a column truncation to eigenvalues greater than some positive numerical tolerance $\epsilon$\footnote{Note that $\epsilon$ is a purely numerical parameter, i.e., in the following, we set it to 0, which suffices if exact algebraic operations are used.} leads to
\begin{align}
    \ullWT{Q}_0 \; \ullWT{\Lambda}_0 \; \ullWT{Q}_0^\mathsf{T} =  \Voigt{\ull{Y}} - \Reuss{\ull{Y}}  + \cO(\epsilon) \, .
\end{align}
From that, we can define a matrix $\ull{L}$ closely related to the Cholesky factorization and its pseudo-inverse~$\ull{L}^+$ via
\begin{align}
    \ull{L} &= \ullWT{Q}_0 \sqrt{\ullWT{\Lambda}_0}, &
    \ull{L}^+ &=  \sqrt{\ullWT{\Lambda}^{-1}_0} \ullWT{Q}^\mathsf{T}_0 \, .
\end{align}
We now modify \cref{eq:preparation} as follows:
\begin{align}
    0 \leq \ullWT{Y} = \ull{L}^+ \lb \Voigt{\ull{Y}} - \ull{Y} \rb {\ull{L}^+}^\mathsf{T} \leq \ull{L}^+ \lb \Voigt{\ull{Y}} - \Reuss{\ull{Y}} \rb {\ull{L}^+}^\mathsf{T} = \ull{I} + \cO(\eps) \, .
\end{align}

Analogous to the one-dimensional case, this reparameterization yields
\begin{align}
    \ull{Y} &= \Voigt{\ull{Y}} - \ull{L} \, \lb \ullWT{Q} \; \ullWT{\Lambda} \; \ullWT{Q}^\mathsf{T} \rb \ull{L}^\mathsf{T}, & \ullWT{Q} & \in Orth(\ffR^{m}), & \ullWT{\Lambda} &= \text{diag} ( {\xi_{\lambda}}_1, \dots , {\xi_{\lambda}}_m), \ {\ul{\xi}_{\lambda}} \in [0, 1]^m \, .
\end{align}
Herein, the term
\begin{align}
    \ullWT{Y} &= \ullWT{Q} \, \ullWT{\Lambda} \; \ullWT{Q}^\mathsf{T} \label{eq:Ytilde}
\end{align}
together with $\Voigt{\ull{Y}}$ and $\Reuss{\ull{Y}}$ suffices to reconstruct $\ull{Y}$.

\subsection{A universal machine-learning model using spectral normalization}
\begin{figure}[!h]
    \centering
    \begin{tikzpicture}[shorten >= 0pt, shorten <= 0pt, rounded corners=4pt]
	\begin{scope}[rectangle, minimum width=2cm, minimum height=1cm, fill=uniSlightblue]
		\node[rectangle, inner sep=0pt, minimum width=2cm, minimum height=1cm, fill=uniSblue, text=white, draw=none, line width=2pt] (input) at (-1.5, 0.5){
			\begin{minipage}{3.25cm}\centering input features $\ul{\chi}$\end{minipage}};
		\node[rectangle, inner sep=0pt, minimum width=2cm, minimum height=1cm, fill=uniSblue, text=white, draw=none, line width=2pt] (trivialinput) at (-1.5, -1.5){
			\begin{minipage}{3.25cm}\centering mandatory: $\ol{\ull{Y}}_\mathrm{V}, \ol{\ull{Y}}_\mathrm{R}$\end{minipage}};
		
		\node[fill=black, text=white, draw=none, inner sep=0pt, minimum height=2cm] (blackbox) at (2,0.5){
			\begin{minipage}{2.5cm}\bfseries\centering black\\box\\model\end{minipage}};
		
		\node[fill=uniSlblue40, draw=none, inner sep=0pt, minimum height=1cm] (Lbox) at (2,-1.5){
			\begin{minipage}{2.5cm}\centering $\ol{\ull{Y}}_\mathrm{V} - \ol{\ull{Y}}_\mathrm{R} \rightarrow \ull{L}$\end{minipage}};
		
		\node[rectangle, minimum width=3cm, inner sep=0pt, minimum height=1cm, fill=uniSlblue40, draw=none, line width=2pt] (out1) at (5.5, 1.125){
			\begin{minipage}{2.95cm}\centering $\ulWH{\xi}_\mathrm{q} \ \rightarrow\ \WH{\ullWT{Q}}$\end{minipage}};
		\node[rectangle, minimum width=3cm, inner sep=0pt, minimum height=1cm, fill=uniSlblue40, draw=none, line width=2pt] (out2) at (5.5, -0.125){
			\begin{minipage}{2.95cm}\centering $\ulWH{\xi}_\lambda \rightarrow \WH{\ullWT{\Lambda}}$\end{minipage}};
		
		\node[rectangle, minimum width=2cm, minimum height=1cm, fill=uniSlblue40, draw=none, line width=2pt] (out3) at (9., 0.5){
			\begin{minipage}{2.25cm}\centering $\WH{\ullWT{Y}} = \WH{\ullWT{Q}}\,\WH{\ullWT{\Lambda}} \, \WH{\ullWT{Q}}^\mathsf{T}$\end{minipage}};
		
		\node[rectangle, minimum width=2cm, inner sep=0pt, minimum height=1cm, fill=red!20, draw=none, line width=2pt] (loss) at (12., 0.5){
			\begin{minipage}{2.cm}\centering loss on $\WH{\ullWT{Y}}$\end{minipage}};
		
		\node[rectangle, minimum width=2cm, minimum height=1cm, text=white, fill=uniSblue, draw=none, line width=2pt] (out4) at (9,-1.5){
			\begin{minipage}{2.75cm}\centering $\WH{\ull{Y}} = \ull{Y}_\mathrm{V} - \ull{L} \, \WH{\ullWT{Y}} \, \ull{L}^\mathsf{T}$\end{minipage}};
		
		\begin{scope}[minimum height=0.75cm, minimum width=3.5cm,inner sep=0pt,yshift=-2.75cm, xshift=2cm]
			\node[draw=none,fill=uniSgray20!60, minimum height=1.1cm, minimum width=14cm, rectangle, rounded corners=4pt,line width=2pt] at (3,0) {};
			\node[rectangle, fill=uniSlblue40] (key1) at (0, 0) {algebraic operations};
			\node[rectangle, fill=uniSblue, text=white] (key2) at (4, 0) {in-/outputs};
			\node[rectangle, fill=red!20] (key3) at (8, 0) {loss};
			\node[draw=none,fill=none, minimum width=0cm] at (-3,0) {\textbf{key:}};
		\end{scope}
		
	\end{scope}
	
	\draw[line width=1pt, -stealth] (input) -- (blackbox);
	\draw[line width=1pt, -stealth] (trivialinput) -- (Lbox);
	\draw[line width=1pt, -stealth] (blackbox) -- (out1);
	\draw[line width=1pt, -stealth] (blackbox) -- (out2);
	\draw[line width=1pt, -stealth] (out1) -- (out3);
	\draw[line width=1pt, -stealth] (out2) -- (out3);
	\draw[line width=1pt, -stealth] (out3) -- (out4);
	\draw[line width=1pt, -stealth] (out3) -- (loss);
	\draw[line width=1pt, -stealth] (Lbox) -- (out4);
\end{tikzpicture}
    \caption{General machine-learning model employing spectral normalization to enforce two-sided Löwner bounds.}
    \label{fig:MLmodel}
\end{figure}
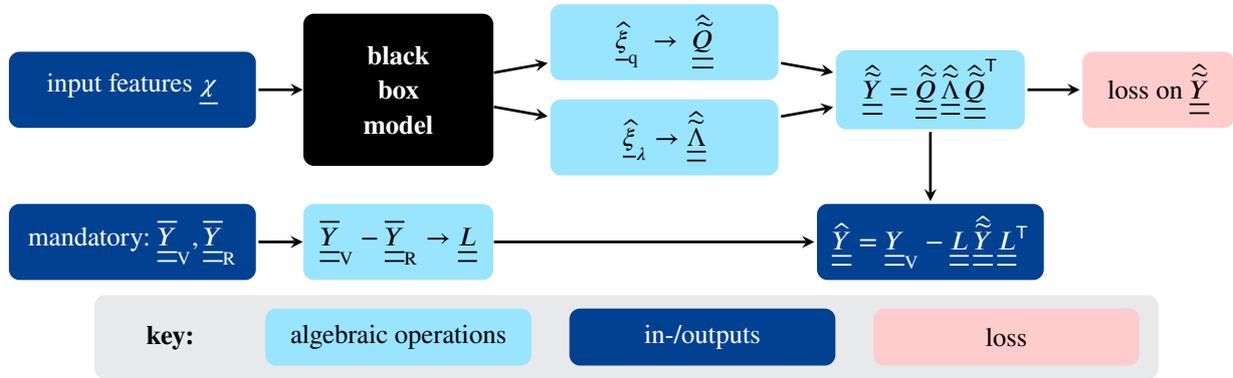

So far, only reformulations of existing quantities have been presented. In this section, we outline how reparameterization through $\ullWT{Q}$, $\ullWT{\Lambda}$ can be combined with machine learning. In the following, a model of a quantity $\bullet$ is denoted $\WH{\bullet}$. Thus, the overall goal is to design an accurate model $\ullWH{Y}$ for the constitutive tensor $\ull{Y}$ given some input features~$\ul{\chi}\in\ffR^{n_\mathrm{in}}$ which, e.g., describe the geometry, material parameters of the individual phases, etc. The above structure helps us to construct such a machine-learned model that strictly respects the upper and lower bounds. Therefore, two distinct contributions are needed (see also \Cref{fig:MLmodel}):
\begin{itemize}
    \item a model $\ullWTH{Q}$ to approximate $\ullWT{Q} \in Orth(\ffR^m)$, via intermediate parameters $\ul{\xi}_\mathrm{q} \in \ffR^{\frac{m(m-1)}{2}} $,
    \item a model $\ullWTH{\Lambda}$ to approximate $\ullWT{\Lambda}$ with $m$ diagonal entries, via intermediate parameters $\ul{\xi}_\lambda \in [0,1]^m$.
\end{itemize}
These two ingredients are represented as a black box model in \Cref{fig:MLmodel}. In particular, the proposed method is independent of the chosen type of model.

The model outputs are,
\begin{align}
    \ullWT{Y} & \approx \ullWTH{Y} = \ullWTH{Q} \, \ullWTH{\Lambda} \, \ullWTH{Q}^\mathsf{T}, &
    \ull{Y} & \approx \ullWH{Y} = \Voigt{\ull{Y}} - \ull{L} \, \ullWTH{Y} \, \ull{L}^\mathsf{T} \, .
\end{align}

Note that the described spectral normalization reduces essentially to a pre-processing step to obtain\footnote{PseudoInverse refers to the Moore-Penrose pseudo inverse \cite{Penrose1955}.}
\begin{align}
\Voigt{\ull{Y}} - \Reuss{\ull{Y}} & = \ull{L} \, \ull{L}^\mathsf{T}, &
\ull{L}^+ &= \text{PseudoInverse}(\ull{L}), &
    \ullWT{Y} &= \ull{L}^+ \lb \Voigt{\ull{Y}} - {\ull{Y}} \rb {\ull{L}^+}^\mathsf{T}
    = \ullWT{Q} \, \ullWT{\Lambda} \, \ullWT{Q}^\mathsf{T} \, .
\end{align}
A relative error with respect to the range of admissible responses $\Voigt{\ull{Y}}-\Reuss{\ull{Y}}$ of an approximation $\ullWH{Y}$ of $\ull{Y}$ can directly be obtained from the absolute error in $\ullWT{Y}$:
\begin{align}
    \phi( \ullWT{Y}, \ullWTH{Y} ) &= \frac{1}{\sqrt{m}}\; \Vert \ullWT{Y} - \ullWTH{Y} \Vert_\mathsf{F} \, , \label{eq:rel:error}
\end{align}
where $\left\| \cdot \right\|_\mathsf{F}$ denotes the Frobenius norm. This loss is devoid of a physical dimension and normalized, i.e., it is insensitive to changes of unit (e.g., SI vs. imperial), and it can be combined with any other de-dimensionalized loss. Therefore, we strongly recommend the use of a loss that is exclusively based on $\ullWTH{Y}$ instead of $\ull{Y}$. Notably, the absolute error is interpretable for each sample $\phi( \ullWT{Y}, \ullWTH{Y} ) \in [0, 1]$. This is because the Frobenius norm of $\ullWT{Y} - \ullWTH{Y}$ is bounded by the sum of the squares of the eigenvalues of $\ullWT{Y} - \ullWTH{Y}$. Interestingly, clever weighting of losses was also shown to be crucial in \cite{Fernandez2021}.

\subsection{Implementation of the spectral normalization}
\label{sec:implementation_spectral_bounds}

As mentioned before, spectral normalization is independent of the type of machine-learning model to be used, i.e., it can be combined with neural networks (in its different flavors), kernel methods, random forests, and many more. The only two essential ingredients are the availability of two utilities:
\begin{itemize}
    \item a vector-valued surrogate $\ulWH{\xi}_\lambda \in [0, 1]^m$ to model $\ullWTH{\Lambda}$ and
    \item a parameterized orthogonal matrix $\ullWTH{Q}\;(\ulWH{\xi}_\mathrm{q})$.
\end{itemize}
The first part is trivial to enforce, e.g., through a simple sigmoid function (or similar alternatives). Parameterization of the orthogonal matrix is more challenging and less commonly available. In our implementation using \texttt{pytorch}~\cite{Ansel_PyTorch_2_Faster_2024}, we deploy \python{torch.nn.parameterizations.orthogonal}\footnote{\url{https://pytorch.org/docs/stable/generated/torch.nn.utils.parametrizations.orthogonal.html}}. In practice, the black box surrogate model must provide $\ulWH{\xi}_\mathrm{q}$, matching the parameterization used.

\begin{remark}
    It is strongly recommended not to compute the loss either based on the parameters $\ul{\xi}_\mathrm{q}$, or on $\ul{\xi}_\lambda$.
    \begin{itemize}
        \item Depending on the chosen parameterization of $\ullWT{Q} \rightarrow \ul{\xi}_\mathrm{q}$, errors in $\ul{\xi}_\mathrm{q}$ are hard to interpret.
        \item Further, redundancy is possible since for any column $\ulWT{q}$ within $\ullWT{Q}$, the use of $-\ulWT{q}$ leads to the same $\ullWT{Y}$.
        \item As for $\ul{\xi}_\lambda$, the ordering of eigenvalues in the surrogate model could be different. The situation is even worse if eigenvalues show algebraic multiplicity, which renders the columns of $\ullWT{Q}$ nonunique.
    \end{itemize} 
    These problems are eliminated when using the suggested loss \cref{eq:rel:error} or any other loss relating $\ullWTH{Y}$ to $\ullWT{Y}$.
\end{remark}

\section{Results}
\label{sec:results}

\subsection{Neural network layout}
\label{sec:nn_layout}

We propose a straightforward feedforward neural network architecture that takes arbitrary microstructure descriptors and phase material parameters as input and outputs a compact representation of $\ullWT{Y} \in \ffR^{m \times m}$ via $\ulWT{y} \in [0,1]^{\frac{m(m+1)}{2}}$. Specifically, for a feedforward neural network with $L$ hidden layers, we write:
\begin{align}
    \ul{z}^{(0)} &= \ul{\chi},  & (\text{input layer}) \\
    \ul{z}^{(\ell)} &= f \;\!\bigl(\ull{W}^{(\ell)} \ul{z}^{(\ell-1)} + \ul{b}^{(\ell)}\bigr), 
    \quad \ell = 1,\dots,L-1,  & (\text{nonlinear hidden layers})\\
    \ulWTH{y} &= \sigma \; \!\Bigl(\ull{W}^{(L)} \ul{z}^{(L-1)} + \ul{b}^{(L)}\Bigr), & (\text{output layer})
\end{align}
where $f(\cdot)$ denotes a suitable pointwise nonlinear activation (e.g., ReLU, Tanh, etc.) and the sigmoid function,
\begin{align}
    \sigma \colon
    \begin{cases}
        \ffR  & \longrightarrow\,\,\, (0,1) \\
      t          & \longmapsto\,\,\, \sigma(t)
    \end{cases}, \qquad \qquad
    &\sigma(t) = \dfrac{1}{1+e^{-t}}, 
\end{align}
which ensures each component of $\ulWH{\xi}_\lambda$ remains in the unit interval $[0,1]$. The weights $\ull{W}^{(\ell)}$ and the biases $\ul{b}^{(\ell)}$, $\ell = 1, \dots, L$ are the training parameters of the neural network that are learned by backpropagation. Notably, the spectral normalization of the effective tensor preserves the intrinsic dimensionality of the learning target as summarized in~\cref{tab:dof_table}.\\

\begin{figure}[htbp]
    \vspace{0.5cm}
    \centering
    \includegraphics[scale=0.25,trim={28cm 6.5cm 15cm 5.9cm},clip]{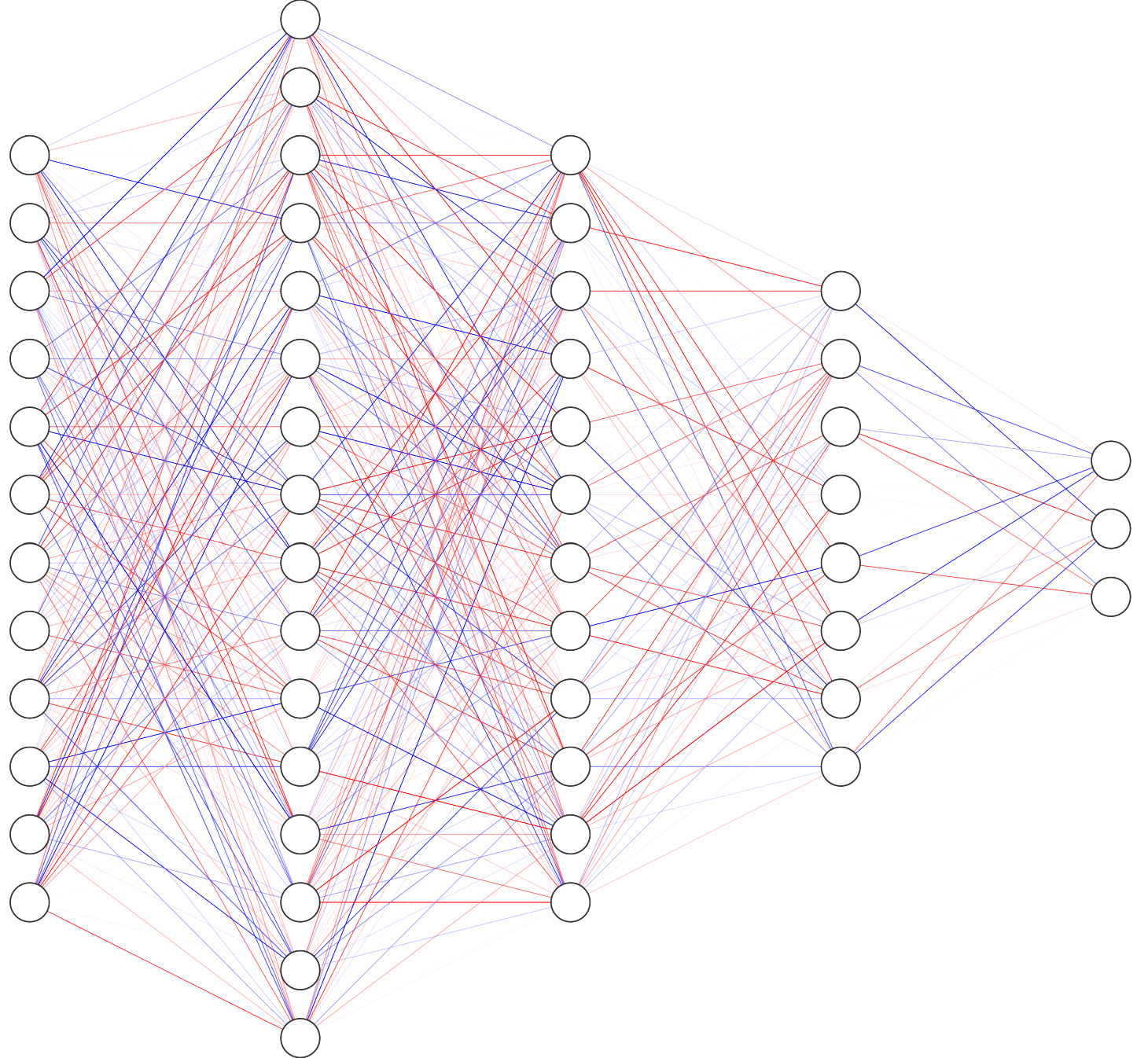}
    \begin{tikzpicture}[overlay, remember picture]
        \node[fill=blue!15, text width=2.5cm, align=center, rotate=90, rounded corners] 
                    at (-7.1, 1.3) {Material parameters};
        \node[fill=red!15, text width=2.5cm, align=center, rotate=90, rounded corners] 
                    at (-7.1, 4.3) {Microstructure descriptors};

        \draw[rounded corners, color=gray, opacity=0.6, dashed] (-5.4, -0.15) rectangle (-1.2, 5.9);
        \node[text width=3.75cm, align=center, rotate=0, rounded corners] 
                    at (-3.3, 6.15) {$L$ nonlinear hidden layers};

        \node[text width=0.8cm, align=center, rotate=0, rounded corners] 
                    at (0.45, 2.8) {$ 
                        \begin{bmatrix} 
                            \; \ulWH{\xi}_\mathrm{q} \;\;\\[0.5em]
                            \; \ulWH{\xi}_\lambda \;\;
                        \end{bmatrix}$};

        \draw[dashed,stealth-stealth,line width=1.5pt] (1.0,2.8) -- node[above] {} (1.9,2.8);

                \node[fill=blue!15, text width=1.2cm, align=center, rotate=0, rounded corners] 
                    at (2.75, 2.8) {$\ullWTH{Y}$};
    \end{tikzpicture}
    \caption{Feed forward neural network architecture for predicting effective property tensor with guaranteed Voigt-Reuss bounds. The feature vector contains the material parameters as well as morphological descriptors representing the microstructure. The output contains normalized eigenvalues $\ulWH{\xi}_\lambda$ and parameterized orthogonal matrix representation $\ulWH{\xi}_\mathrm{q}$ that maps to a physically valid effective tensor $\ullWH{Y}$ via $\ullWTH{Y}$.}
    \label{fig:nn_architecture}
\end{figure}
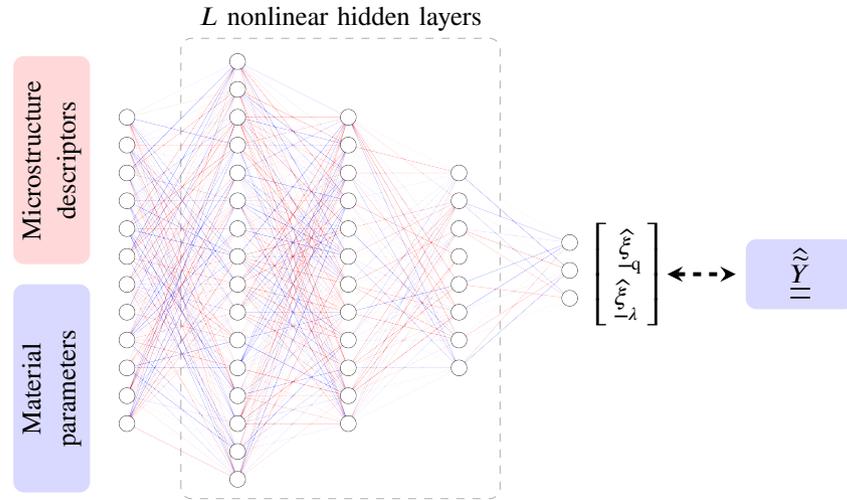

Since network outputs are normalized degrees of freedom for the effective property tensor, the input layer $\ul{z}^{(0)}$ can accommodate any feature vector describing the microstructure (e.g., morphological descriptors or other domain-specific attributes). This design choice makes the approach agnostic to input representation, and the spectral normalization is applied purely to the output using the Voigt and Reuss bounds.

\begin{table}[t]
\centering
\caption{Comparison of unnormalized vs.\ normalized degrees of freedom for 2D/3D thermal problems. Here, degrees of freedom (DOF) refers to the intrinsic target dimension of the ML model}
\label{tab:dof_table}
\begin{tabular}{lccccc}
\toprule
\multirow{2}{*}{\textbf{Problem}} & 
\multirow{2}{*}{\textbf{Dimension}} & 
\multicolumn{2}{c}{\textbf{Unnormalized DOFs}} & 
\multicolumn{2}{c}{\textbf{Normalized DOFs}} 
\\
\cmidrule(lr){3-4}\cmidrule(lr){5-6}
 & & Matrix form & \# DOFs & Eigenvals + Orth.\ DOFs & \# DOFs \\
\midrule
Thermal & 2D & $2 \times 2$ sym.\ ($\ol{\ull{\kappa}}$) & $3$ 
        & $\ul{\xi}_\mathrm{q}\in \ffR^{1}$,\;\; $\ul{\xi}_\lambda \in \ffR^{2}$  & $3$ \\[2.5pt]
Thermal & 3D & $3 \times 3$ sym.\ ($\ol{\ull{\kappa}}$) & $6$ 
        & $\ul{\xi}_\mathrm{q} \in \ffR^{3}$,\;\; $\ul{\xi}_\lambda \in \ffR^{3}$ & $6$ \\
\bottomrule
\end{tabular}
\end{table}

\subsection{High-fidelity data generation via Fourier-Accelerated Nodal Solvers (FANS)}
\label{subsec:fans_intro}

In this section, the methodology for high-fidelity numerical simulations that serve as training, validation, and test data for the surrogate models is presented. These simulations were carried out using the \emph{Fourier-Accelerated Nodal Solvers} (FANS)\cite{Leuschner2017}. Our in-house developed FANS code~\footnote{\url{https://github.com/DataAnalyticsEngineering/FANS}} is an open-source, parallel, high-performance C++ implementation of FANS for microscale multiphysics problems. FANS is a specific Fast Fourier Transform (FFT) --based technique for the numerical homogenization of heterogeneous microstructures for linear and nonlinear multi-physical problems. FANS operate directly on high-resolution voxel-based images and exploit the FFT to perform repeated periodic convolutions in Fourier space to accelerate convergence. For high-performance scalability, the code exploits an efficient parallel FFT library- Fast Fourier transform in the west \texttt{FFTW}---and parallelism via Message passing interface \texttt{MPI}. This ensures that even large 3D microstructures (up to hundreds or thousands of voxels in each dimension) can be simulated in a reasonable time and memory footprint. 

\subsection{Application to 2D linear thermal homogenization}
\label{subsec:2d_thermal_homogenization}

In this section, we present the application of the spectral normalization to predict the microstructure-dependent 2D linear thermal conductivity tensor ~$\ol{\fkappa}$ (see \Cref{subsec:3d_thermal_homogenization} for 3D results). The biphasic materials are both isotropic with conductivity $\kappa_1$ and $\kappa_2$, respectively. The effective thermal conductivity $\ol{\fkappa}$ is the quantity of interest extracted from the microstructural data. It is represented as $2\times 2$ symmetric positive definite matrix $\ol{\ull{\kappa}}$, which replaces $\ull{Y}$ in the general description in \Cref{sec:spectral:normalization}.

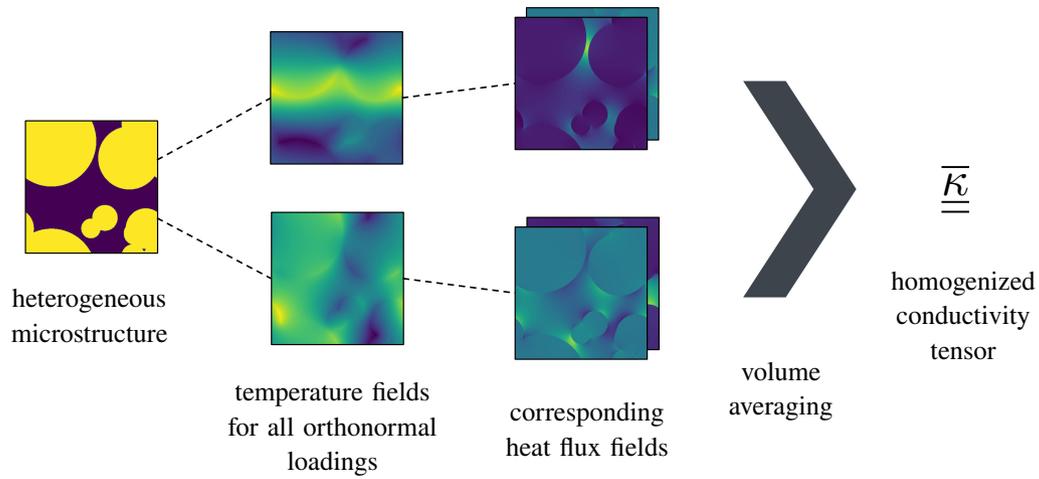
\begin{figure}
    \centering
    \includegraphics[page=4,scale=0.6,trim={2cm 2.5cm 2cm 3cm},clip]{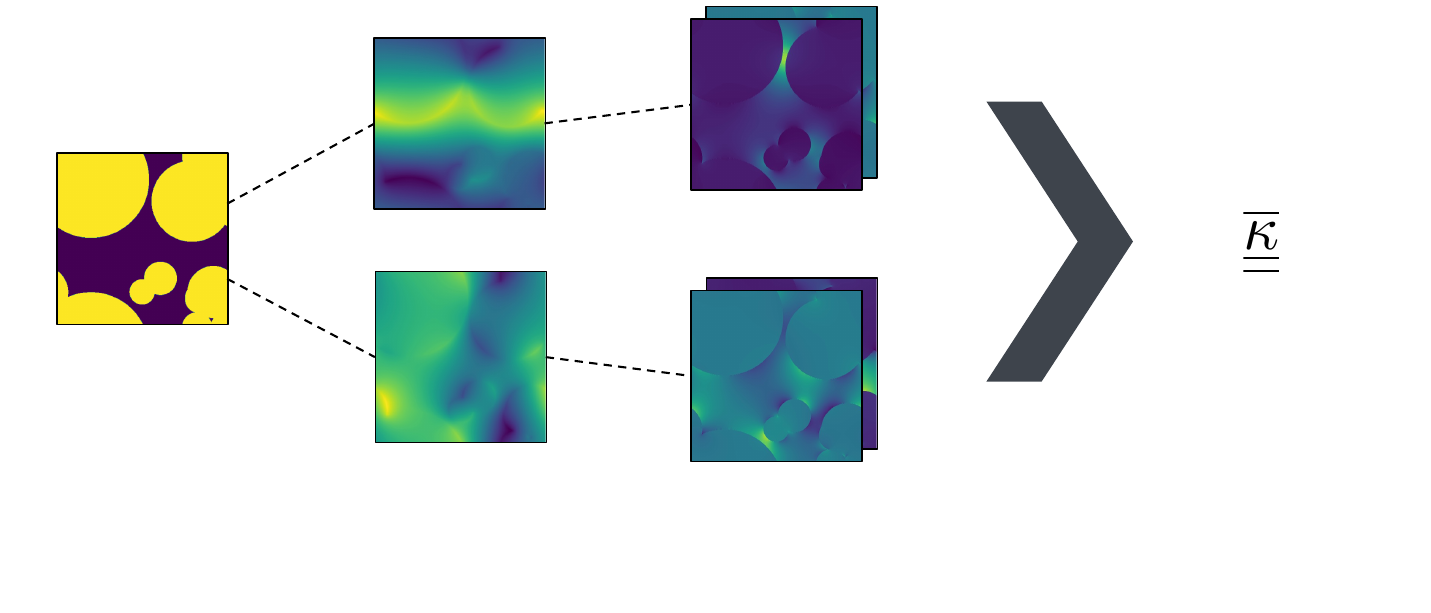}
    \caption{Workflow for constructing the effective conductivity tensor $\ol{\fkappa}$ from high-fidelity simulations.}
    \begin{tikzpicture}[overlay, remember picture]
        \node[fill=white, text width=2.5cm, align=center] at (-5.85, 3.25) {heterogeneous microstructure};
        \node[fill=white, text width=3.75cm, align=center] at (-2.65, 1.75) {temperature fields for all orthonormal\\ loadings};
        \node[fill=white, text width=3.0cm, align=center] at (0.7, 1.75) {corresponding heat flux fields};
        \node[fill=white, text width=1.5cm, align=center] at (3.25, 2.25) {volume averaging};
        \node[fill=white, text width=2.65cm, align=center] at (5.65, 3.25) {homogenized\\ conductivity tensor};
    \end{tikzpicture}
    
    \label{fig:orthonormal_loading_to_kappa}
\end{figure}

\subsubsection{2D microstructures and descriptors}
\label{subsubsec:2d_microstructures}

We first focus on a collection of 2D periodic biphasic microstructures that serve as the foundation for training and validating the surrogate model. These periodic microstructure images were generated at a resolution of $400^2$ pixels by a \emph{random sequential adsorption} algorithm, as detailed in~\cite{lissner2019}. A diverse set of morphological features was employed to create 15{,}000 microstructure images, each with circular and rectangular inclusions (training), and 1{,}500 microstructure images with mixed elliptical and rectangular inclusions (validation). 

\begin{figure}[ht]
    \centering
    \includegraphics[page=1, scale=1.15, trim={0.0cm 0.0cm 0.0cm 0.0cm}, clip]{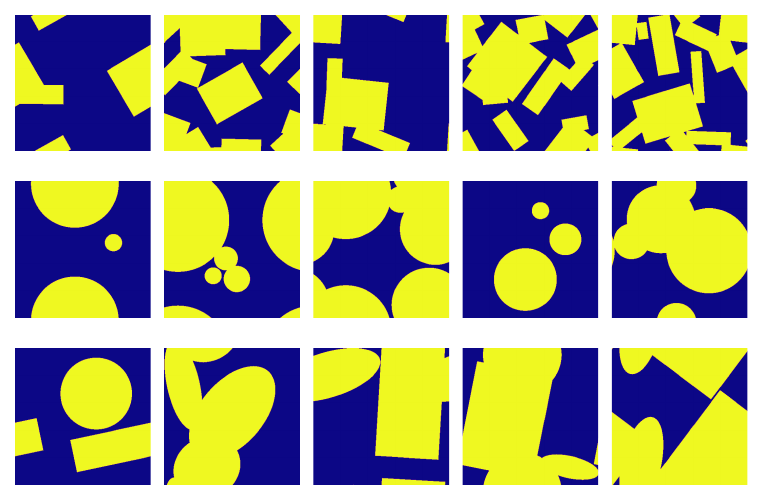}
    \caption{Some exemplary 2D microstructures used for training and validation.}
    \begin{tikzpicture}[overlay, remember picture]
        \node[fill=gray!20, text width=2.55cm, align=center, rotate=90, rounded corners] at (-8, 2.6) {\textbf{Validation set}};
        \node[fill=gray!20, text width=5.75cm, align=center, rotate=90, rounded corners] at (-8, 7.55) {\textbf{Training set}};
        \node[fill=white, text width=2.5cm, align=center, rotate=90] at (-7.4, 2.7) {Mixed};
        \node[fill=white, text width=2.65cm, align=center, rotate=90] at (-7.4, 5.75) {Circular};
        \node[fill=white, text width=3.15cm, align=center, rotate=90] at (-7.4, 9.0) {Rectangular};
    \end{tikzpicture}
    \label{fig:example_2D_microstructures}
\end{figure}
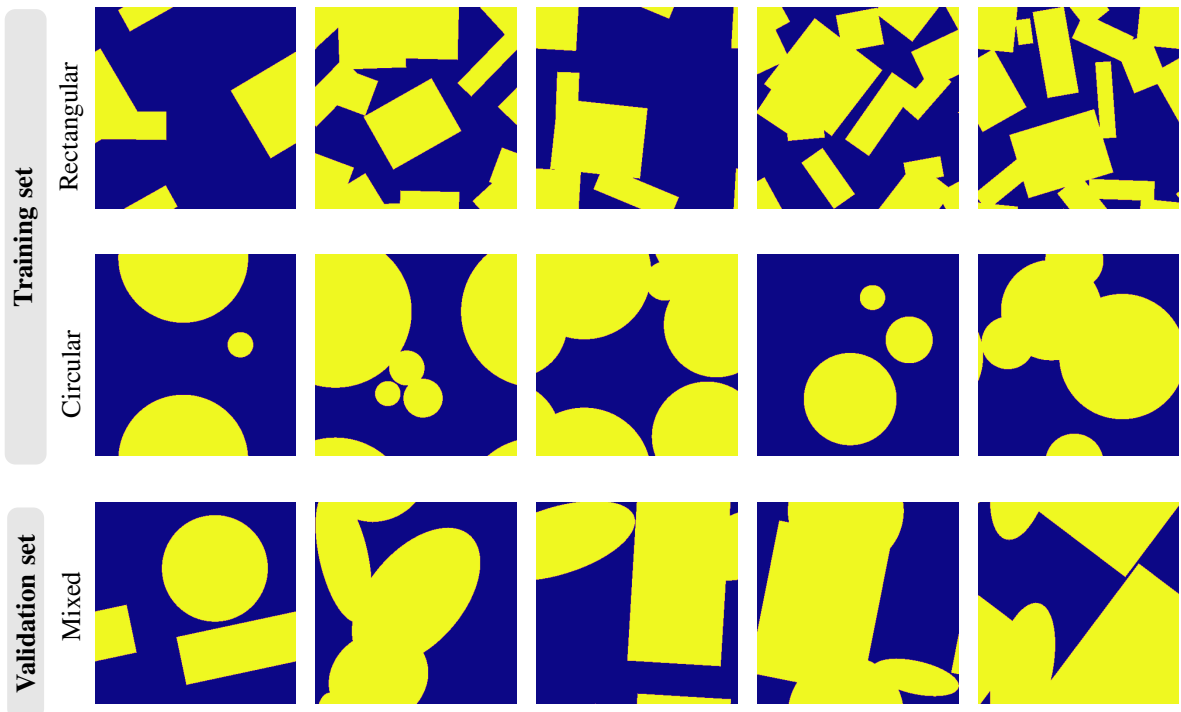

The microstructure images were generated with a wide phase volume fraction range (18 to 84\%), inclusion positions and sizes relative to the image dimensions, and an overlap parameter to control the degree of intersection between inclusions. The orientation and aspect ratio of the rectangular inclusions were also randomized. This ensures a highly varied suite of microstructures spanning different degrees of phase connectivity, inclusion anisotropy, etc. Samples of the resulting images are shown~\cref{fig:example_2D_microstructures}, illustrating the broad geometric variability captured by this dataset. Notably, the validation set lies outside the strict morphological family used for training, which presents a challenging test for the surrogate’s ability to generalize beyond the original geometry classes.

Rather than passing the microstructure images directly to the neural network, each microstructure is characterized by 51 geometric descriptors designed to capture essential morphological features.
These descriptors include
\begin{itemize}
    \item \textbf{Volume fraction} of inclusions (1 scalar).
    \item \textbf{Reduced-basis coefficients} of the two-point correlation function (13 scalars).
    \item \textbf{Band features} to quantify phase connectivity along specific directions (16 scalars).
    \item \textbf{Global directional means} to capture flux hindrance across principal axes (2 scalars).
    \item \textbf{Volume fraction distribution} to reflect different scales of inclusion size (7 scalars).
    \item \textbf{Directional edge distribution} to encode the shape and orientation of inclusions (12 scalars).
\end{itemize}
Details of these features can be found in~\cite{lissner2024}. The resulting descriptors provide a compact yet descriptive representation that is well-suited for training data-driven models while greatly reducing computational overhead. All microstructure images and related descriptors are publicly available as part of an open-access dataset \cite{lissner2023_dataset}.

\subsubsection{2D Linear thermal homogenization surrogate}

The data on which we are building this study are obtained from high-fidelity FANS simulations as described in~\cref{subsec:fans_intro} on synthetic biphasic microstructures described in~\cref{subsubsec:2d_microstructures}. The phase contrast~$R$ is defined as
\begin{align}
    R = \dfrac{\kappa_1}{\kappa_2},
\end{align}
where $\kappa_1$ and $\kappa_2$ are the isotropic conductivities of the two phases (1: matrix, 2: inclusion). For data generation, we fix $\kappa_1 = 1$ and vary $\kappa_2$ to obtain a set of phase contrasts $R$ that span several orders of magnitude. Due to the linearity of the governing equations, any scaling of $\kappa_1$ can be applied \textit{a posteriori}, thereby ruling out one redundant parameter. The training and validation data sets are constructed as detailed in~\cref{tab:phase_contrast_sims2D}.

\begin{table}[htbp]
    \centering
    \caption{Summary of the datasets considered for the 2D thermal homogenization problem.}
    \label{tab:phase_contrast_sims2D}
    \begin{tabular}{lccc}
        \toprule
        \textbf{Dataset} & \textbf{Phase contrast} $R$ & \textbf{Microstructures} & \textbf{Samples} \\[0.1cm]
        \midrule
        Train   & $\left\{\dfrac{1}{100},\;\dfrac{1}{50},\;\dfrac{1}{20},\;\dfrac{1}{10},\;\dfrac{1}{5},\;\dfrac{1}{2},\;2,\;5,\;10,\;20,\;50,\;100\right\}$ 
                & 30,000 
                & 360,000 \\[0.3cm]
        Validation & $\left\{\dfrac{1}{100},\;\dfrac{1}{99},\;\dfrac{1}{98},\ldots,\dfrac{1}{3},\;\dfrac{1}{2},\;2,\;3,\ldots,98,\;99,\;100\right\}$ 
                & 1,500 
                & 297,000 \\[0.2cm]
        \bottomrule \\[-0.1cm]
        \textbf{Total} & & \textbf{31,500} & \textbf{657,000} \\
        \bottomrule
    \end{tabular}
\end{table}

The validation dataset is built from a distinctly different class of microstructures: 1,500 images featuring mixed elliptical and rectangular inclusions, while the training set comprises circular and rectangular inclusions (see~\cref{fig:example_2D_microstructures}). Thus, the validation set contains not only microstructures that are outside the training morphology but also material parameters that were never encountered during training, thereby presenting a stringent test of generalization.

\begin{figure}[htbp]
    \centering
        \includegraphics[scale=1.0,trim={0cm 0cm 0cm 0cm},clip]{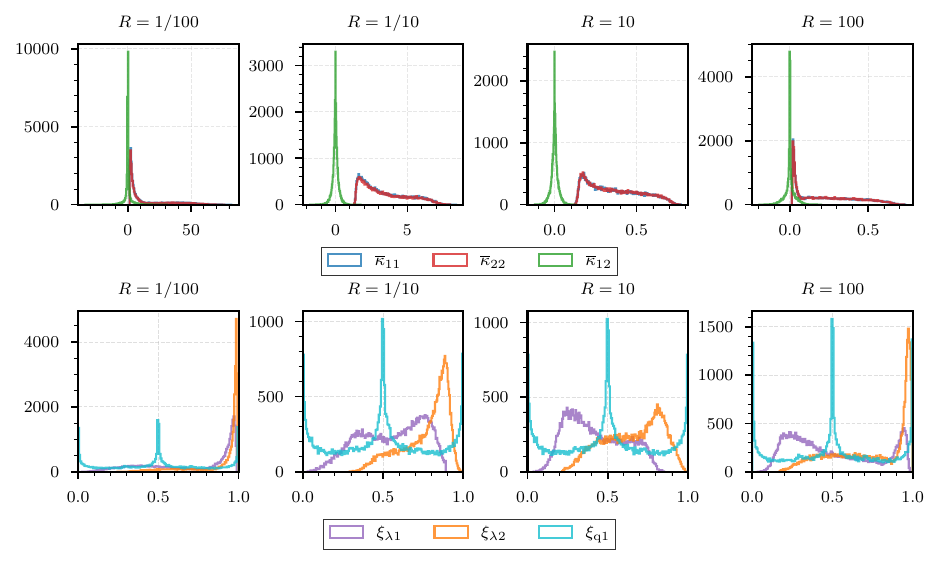}
    \begin{tikzpicture}[overlay, remember picture]
        \node[fill=gray!20, text width=3.25cm, align=center, rotate=90, rounded corners] at (-16.2, 7.4) {Unnormalized targets};
        \node[fill=gray!20, text width=3.25cm, align=center, rotate=90, rounded corners] at (-16.2, 2.8) {Normalized targets};
    \end{tikzpicture}
    \caption{Histograms of the unnormalized ($\ol{\kappa}_{11}$, $\ol{\kappa}_{22}$ and $\ol{\kappa}_{12}$) and the normalized ($\xi_{\lambda1}$, $\xi_{\lambda2}$ and $\xi_{\mathrm{q}1}$) training set targets for varying phase contrasts.}
    \label{fig:therm2d_histograms} 
\end{figure}
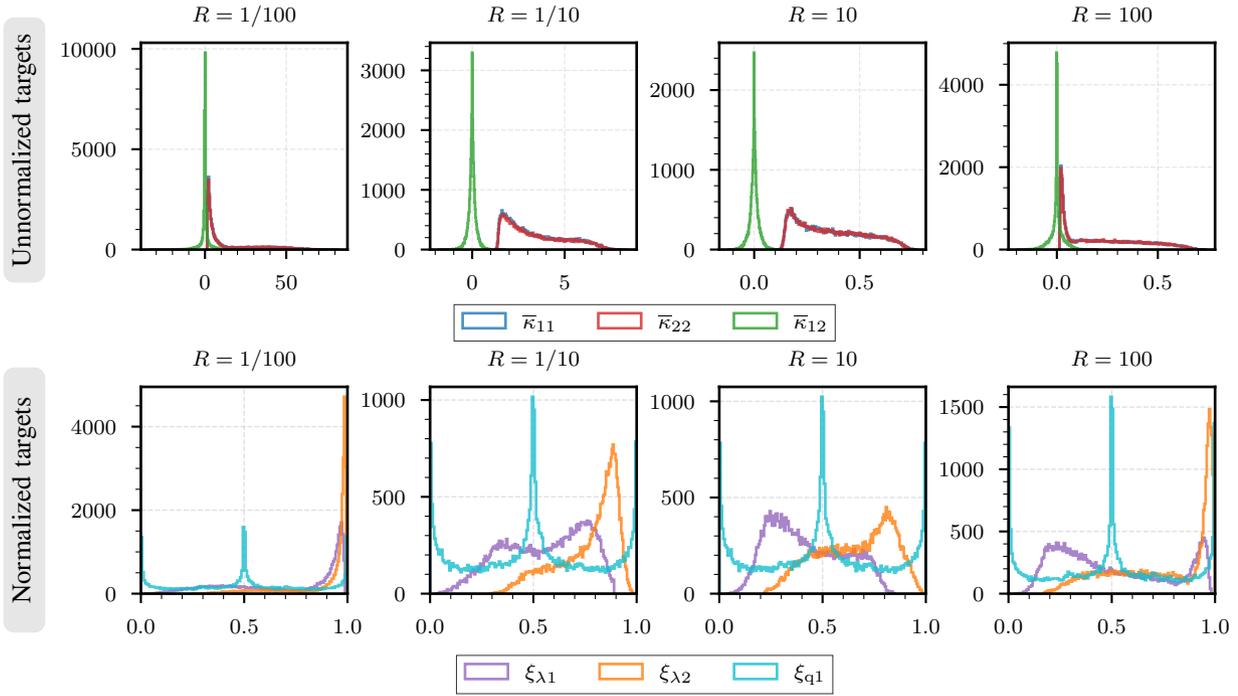

The target quantities are the components of the effective conductivity tensor $\ull{\ol{\kappa}}$, namely $\ol{\kappa}_{11}$, $\ol{\kappa}_{22}$, and $\ol{\kappa}_{12}$. Due to the highly complex influence of phase contrast, the range of these targets is unbounded and varies dramatically with $R$. The~\cref{fig:therm2d_histograms} displays histograms of the unnormalized tensor components for four representative phase contrasts ($R=1/100$, $R=1/10$, $R=10$, and $R=100$). In contrast, spectral normalization maps these values into a bounded interval and the corresponding histograms for the normalized targets: eigenvalues $\xi_{\lambda1}$, $\xi_{\lambda2}$, and a single orthogonal degree of freedom $\xi_{\mathrm{q}1}$, all of which are strictly confined to the interval $[0,1]$ in the parameterization used in our implementation. This uniform range is expected to simplify the training process.

\begin{figure}[htbp]
    \centering
    \begin{subfigure}{0.495\textwidth}
        \centering
        \includegraphics[scale=1.0]{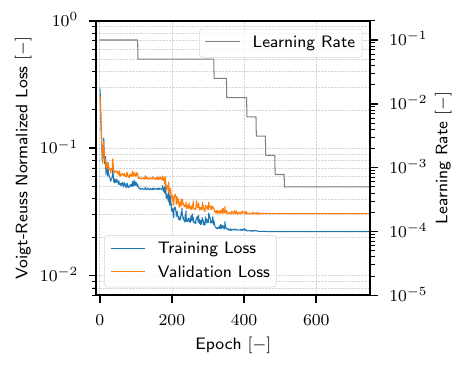}
        \label{fig:therm2d_vrnn_training_history}
    \end{subfigure}
    \hfill
    \begin{subfigure}{0.495\textwidth}
        \centering
        \includegraphics[scale=1.0]{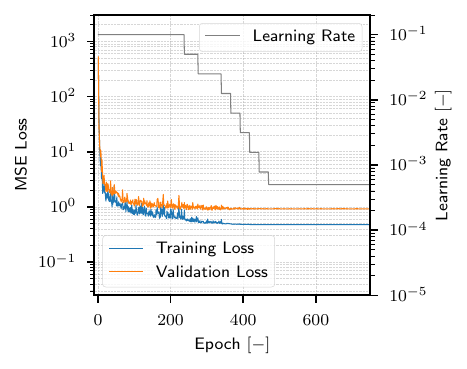}
        \label{fig:therm2d_vann_training_history}
    \end{subfigure}
    \caption{Training history of an exemplary training run for the \VRNet{} (left) and the vanilla neural network (right) for the 2D heat conduction problem. Both networks have the same dense architecture (number of neurons: $[256,\,128,\,64,\,32,\,16]$, activation functions in each layer: \texttt{[SELU, TANH, SIGMOID, Identity]} each applied in one-fourth of the neurons. }
    \label{fig:therm2d_training_history}
\end{figure}

Two candidate neural network models are considered:
\begin{itemize}
    \item A \textit{vanilla neural network} that is trained to predict the unnormalized components $\ol{\kappa}_{11}$, $\ol{\kappa}_{22}$, and $\ol{\kappa}_{12}$.
    \item The proposed \VRNet{} is instead trained to predict normalized targets $\ulWT{y}$ (that is, $\ul{\xi}_\lambda$ and $\ul{\xi}_\mathrm{q}$), which are then back-transformed to recover $\ull{\ol{\kappa}}$.
\end{itemize}
Both models share the same input features (51 geometric descriptors), architecture, and training protocol. The dimensions of the hidden layer are set to $[256,\,128,\,64,\,32,\,16]$ with an output layer of size 3, resulting in a total of 58,627 parameters (including batch normalization). Concerning the activation functions, we deviate slightly from the classical approach that would use the same activation on all neurons. Instead, we partition the neurons into 4 equally sized batches on each layer and apply the following activations: \texttt{[SELU, TANH, SIGMOID, Identity]}. The identity function acts as a bypass, which can also help propagate gradient information through the layers. The vanilla model and the \VRNet{} model are both trained for 750 epochs using the \texttt{ADAM}~\cite{Kingma2014AdamAM} optimizer and a \texttt{ReduceLROnPlateau}\footnote{\url{https://pytorch.org/docs/stable/generated/torch.optim.lr_scheduler.ReduceLROnPlateau.html}} learning rate scheduler. The~\cref{fig:therm2d_training_history} shows the training and validation losses, as well as the evolution of the learning rate for an exemplary training run, for the \VRNet{} and the vanilla network. Both losses stagnated after 400 epochs for either of the two models, hinting at a converged set of parameters.

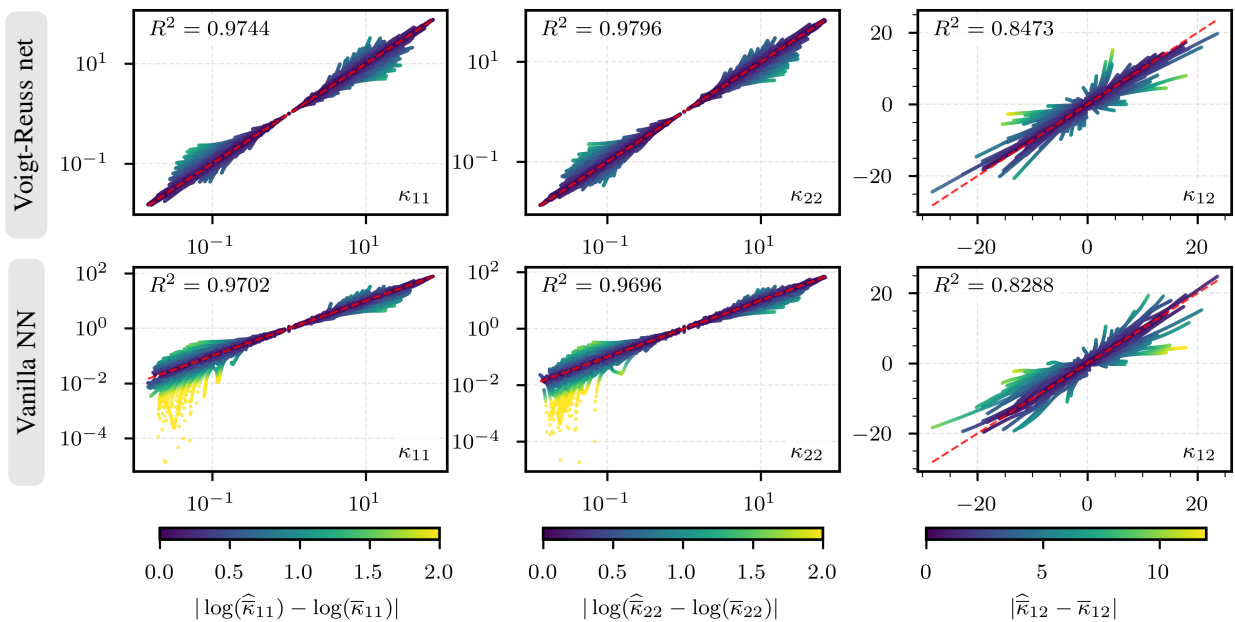
\begin{figure}[htbp]
    \centering
    \includegraphics[scale=1.0]{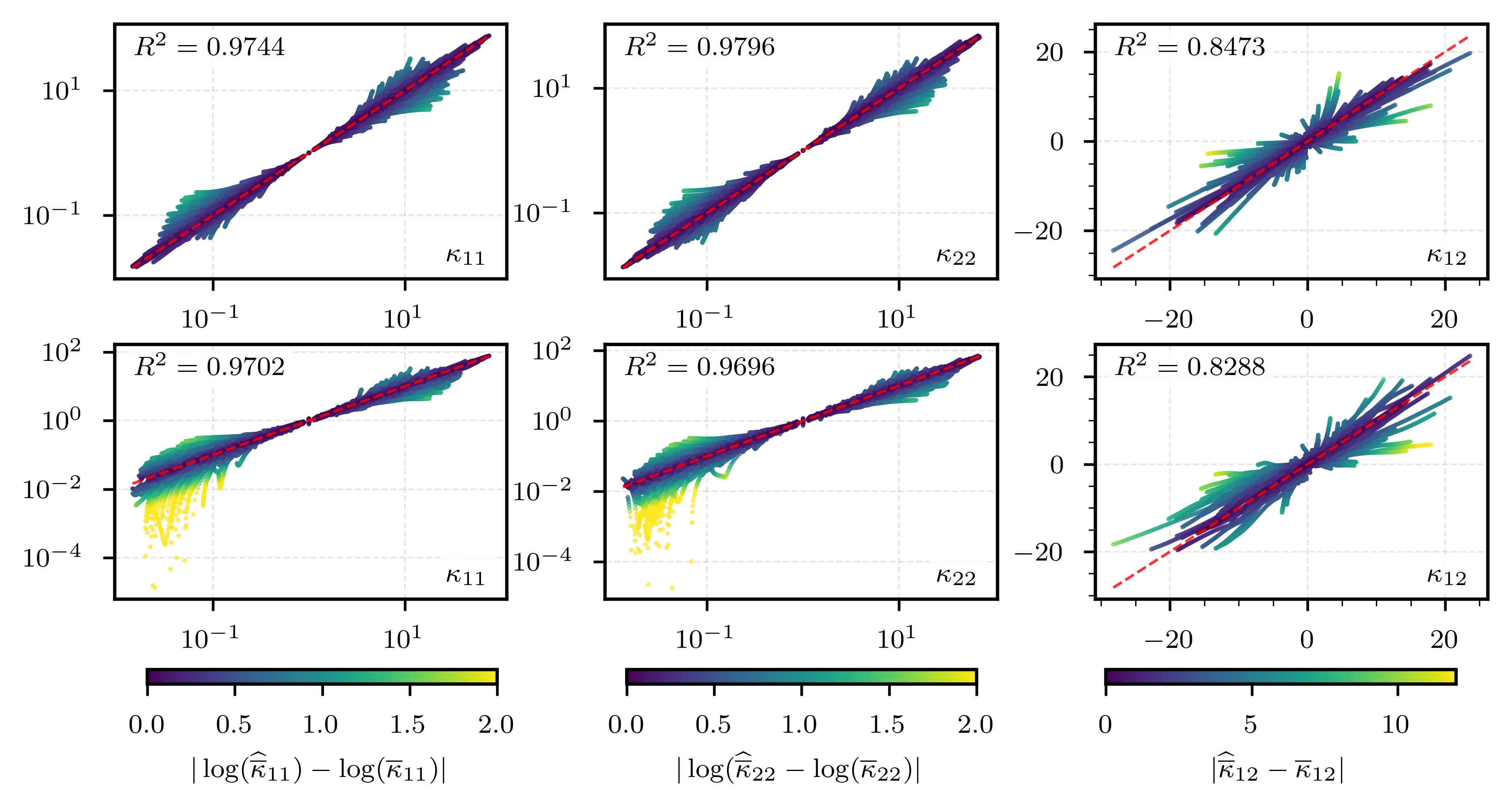}
    \begin{tikzpicture}[overlay, remember picture]
        \node[fill=gray!20, text width=2.75cm, align=center, rotate=90, rounded corners] at (-16.2,6.75) {\VRNet{}};
        \node[fill=gray!20, text width=2.75cm, align=center, rotate=90, rounded corners] at (-16.2,3.5) {Vanilla NN};
    \end{tikzpicture} 
    \caption{Validation data predictions of the effective conductivity tensor components for the \VRNet{} (top) and the vanilla network (bottom) for the 2D heat conduction problem.}
    \label{fig:therm2d_predictions_comparision}
\end{figure}

The predictive capabilities of the models considered are demonstrated in~\cref{fig:therm2d_predictions_comparision}. In the upper row, the results from the \VRNet{} are shown; here, the normalized output of the network is transformed back into the physical components $\ol{\kappa}_{11}$, $\ol{\kappa}_{22}$, and $\ol{\kappa}_{12}$. The predictions of all the validation data are considered and are highly clustered around the identity line, which is reflected by elevated $R^2$ scores and a marked absence of extreme outliers. In contrast, the lower row, which details the predictions from the vanilla network, exhibits considerably more scatter and several anomalous predictions that deviate substantially from the expected trend.

\begin{figure}[htbp]
    \centering
    \includegraphics[scale=1.0]{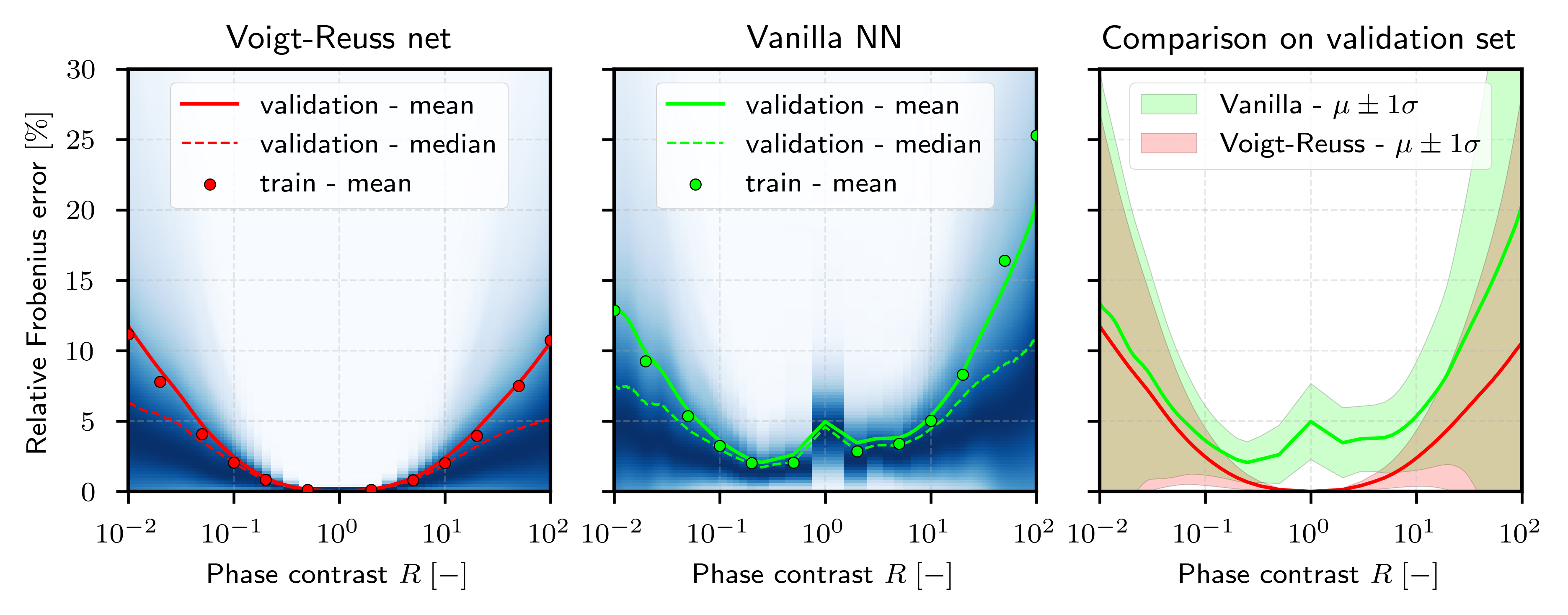}
    \caption{Error distribution of the \VRNet{} and the vanilla network for varying phase contrasts in the 2D heat conduction problem.}
    \label{fig:2D_thermal_error_comparison}
\end{figure}

Further information is provided by~\cref{fig:2D_thermal_error_comparison}, where a comprehensive error analysis is presented across various phase contrasts. The relative Frobenius error on all validation data for the \VRNet{} is shown as a function of phase contrast $R$, highlighting mean and median errors and also incorporating a kernel density estimate that captures the overall distribution of errors. A similar set of error statistics for the vanilla network is presented. A direct comparison illustrating the mean errors with a one-standard deviation ($1\sigma$) band for both models is also shown. This indicates that the \VRNet{} consistently produces lower error statistics across the spectrum of phase contrasts, and the smooth variation in its error profile indicates a more robust interpolation ability for material properties. Additionally, the mean errors for training and validation data are of striking similarity for the \VRNet{} even at extreme phase contrasts, while the vanilla model showed some notable deviations for $R \to 100$.

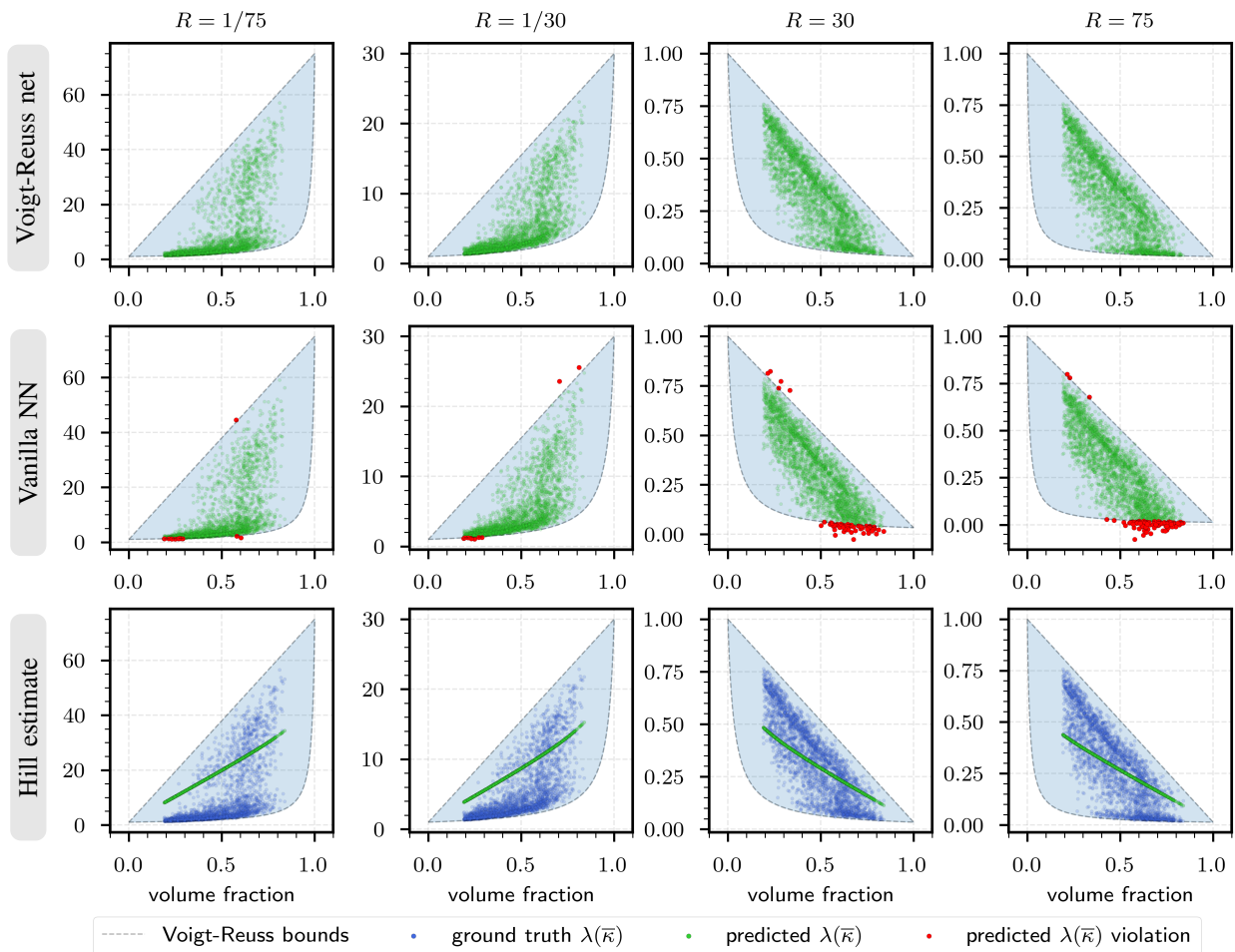
\begin{figure}[htbp]
    \centering
    \includegraphics[scale=1.0]{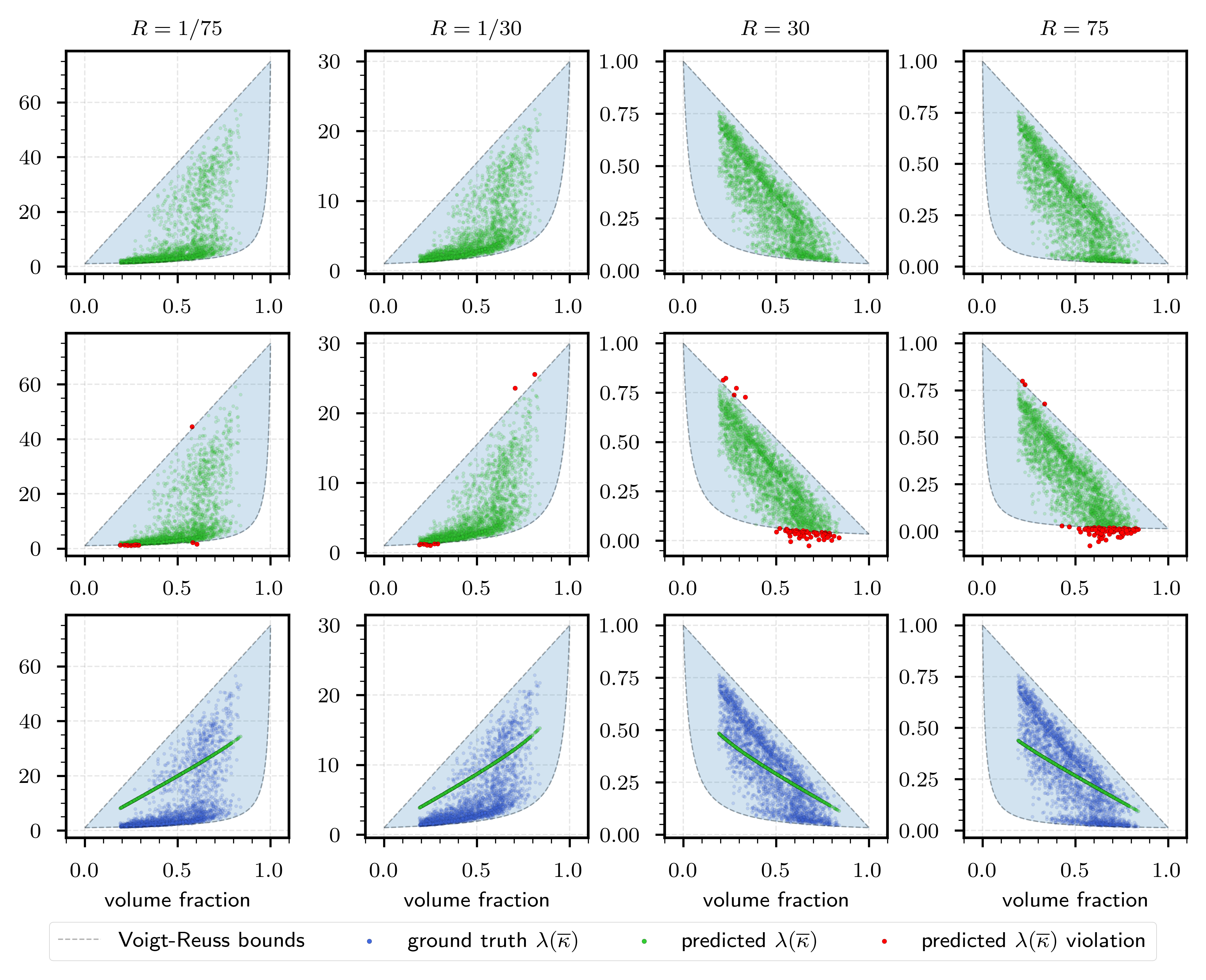}
    \begin{tikzpicture}[overlay, remember picture]
        \node[fill=gray!20, text width=2.75cm, align=center, rotate=90, rounded corners] at (-16.2,10.75) {\VRNet{}};
        \node[fill=gray!20, text width=2.75cm, align=center, rotate=90, rounded corners] at (-16.2,7.0) {Vanilla NN};
        \node[fill=gray!20, text width=2.75cm, align=center, rotate=90, rounded corners] at (-16.2,3.25) {Hill estimate};
    \end{tikzpicture} 
    \caption{Spectral ground truth and predictions of the effective conductivity tensor for varying phase contrasts in the 2D heat conduction problem for all of the validation microstructures.}
    \label{fig:therm2d_predictions_VRbounds}
\end{figure}

A key feature of the \VRNet{} is its capability to enforce the theoretical bounds on the effective conductivity tensor. In~\cref{fig:therm2d_predictions_VRbounds}, a detailed illustration using four selected exemplary phase contrasts from the validation set: $R = 1/75$, $R = 1/30$, $R = 30$, and $R = 75$ is presented. For each contrast, the figure shows the eigenvalues of the predicted $\ull{\ol{\kappa}}$ by both the \VRNet{} and the vanilla network, along with the eigenvalues of the ground truth and the Voigt-Reuss (Hill) average. The \VRNet{} consistently yields predictions that remain strictly within the prescribed bounds, highlighting its intrinsic physical consistency. In contrast, the vanilla network sometimes deviates beyond these limits. Although the Hill estimate is always admissible, its predictive accuracy does not match that of the \VRNet{}. This comparison emphasizes the effectiveness of spectral normalization in ensuring physical plausibility and delivering superior accuracy across a wide range of phase contrasts. These results collectively demonstrate that the spectral normalization, when embedded within a surrogate model, can significantly enhance both prediction accuracy and physical consistency in 2D linear thermal homogenization problems.

\subsection{Application to 3D linear thermal homogenization}
\label{subsec:3d_thermal_homogenization}

\subsubsection{3D microstructures and descriptors}
\label{subsubsec:3d_microstructures}

\begin{figure}[h!]
    \centering
    \begin{tabular}{cccc}
        \subcaptionbox{Fiber system                         \label{fig:subfig1}}{\includegraphics[width=0.235\textwidth,trim={17cm 0 17cm 3cm},clip]{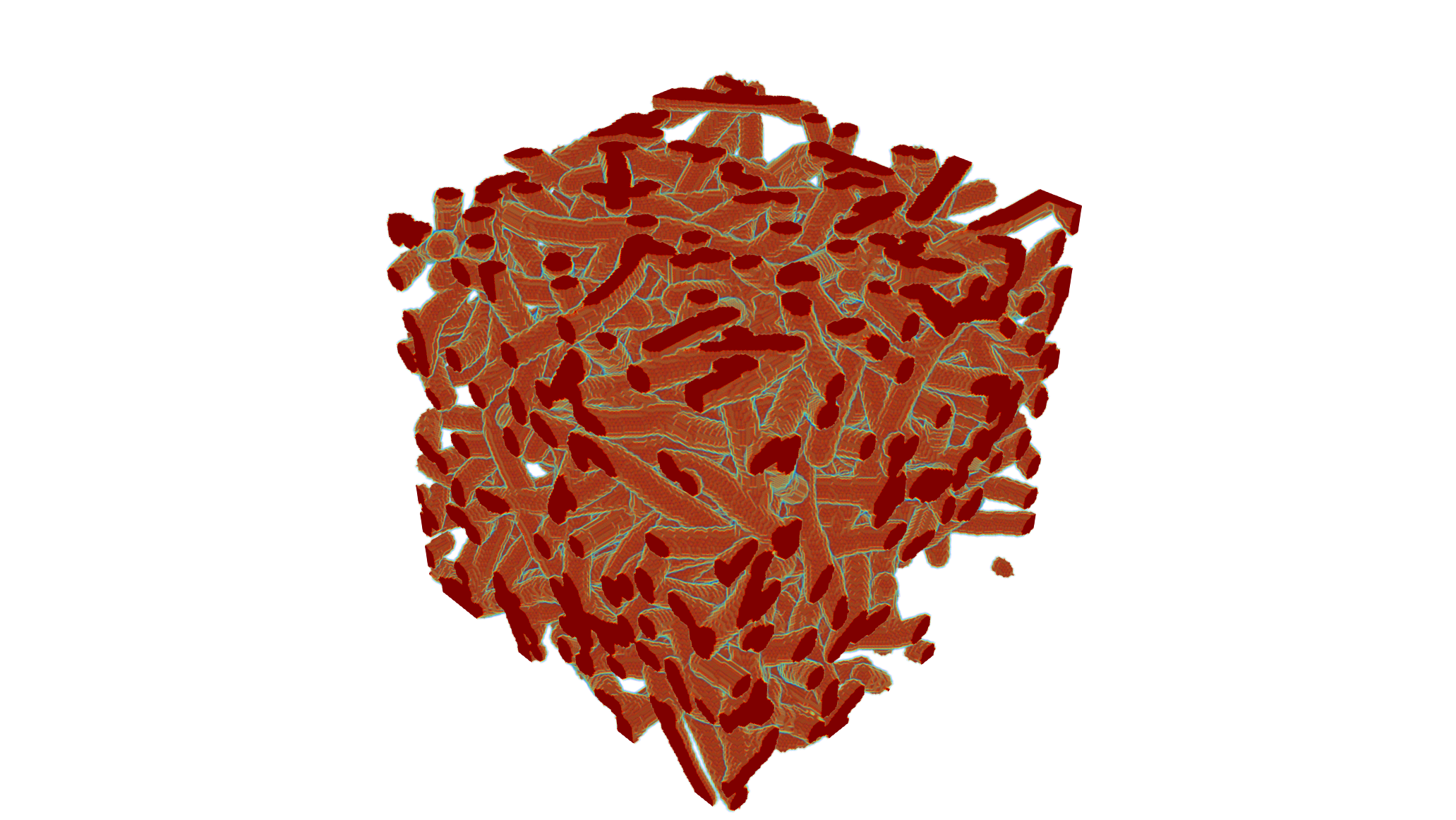}} & 
        \subcaptionbox{Channel system                       \label{fig:subfig2}}{\includegraphics[width=0.235\textwidth,trim={17cm 0 17cm 3cm},clip]{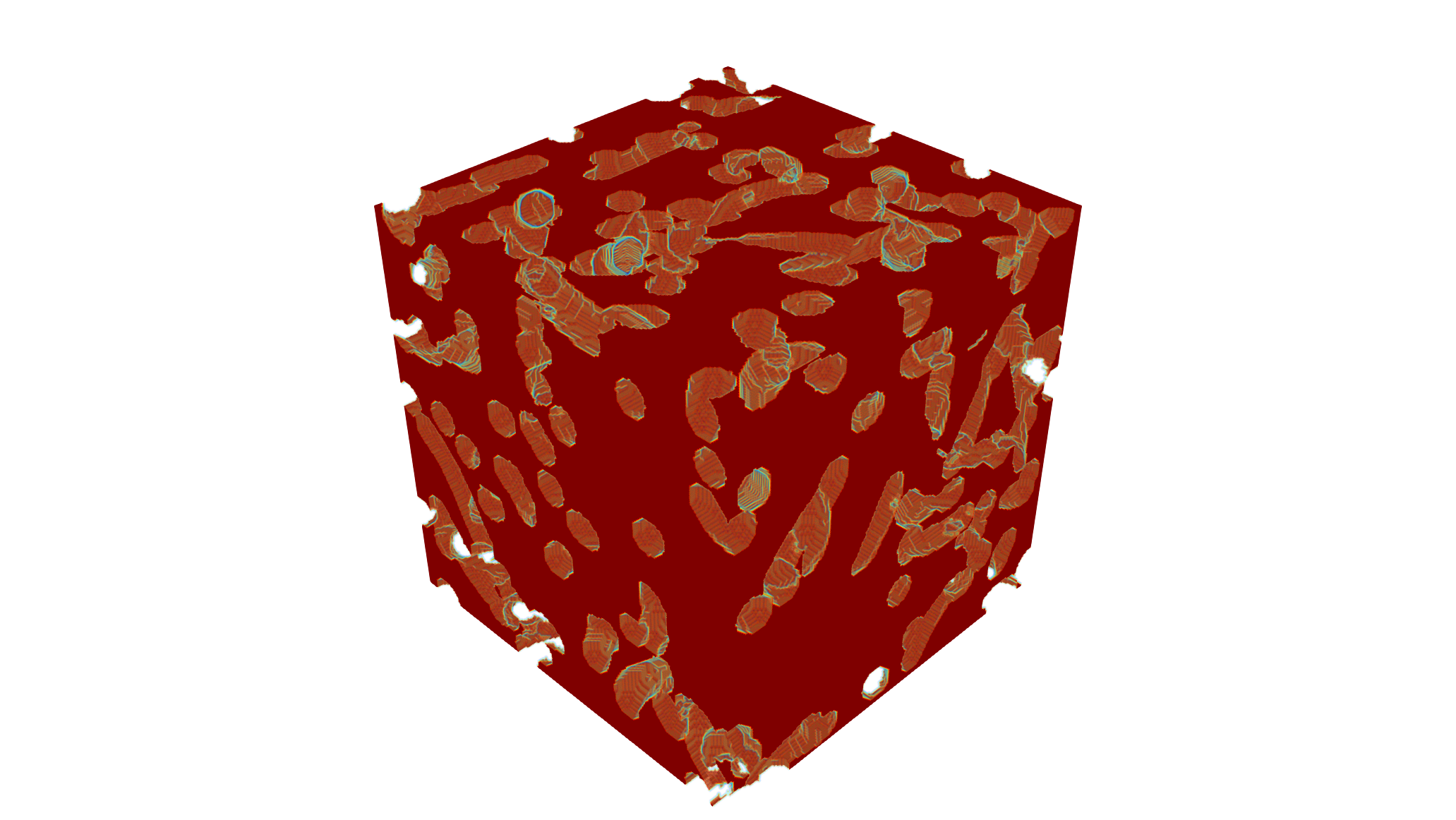}} & 
        \subcaptionbox{Level set of Gaussian random fields  \label{fig:subfig4}}{\includegraphics[width=0.235\textwidth,trim={17cm 0 17cm 3cm},clip]{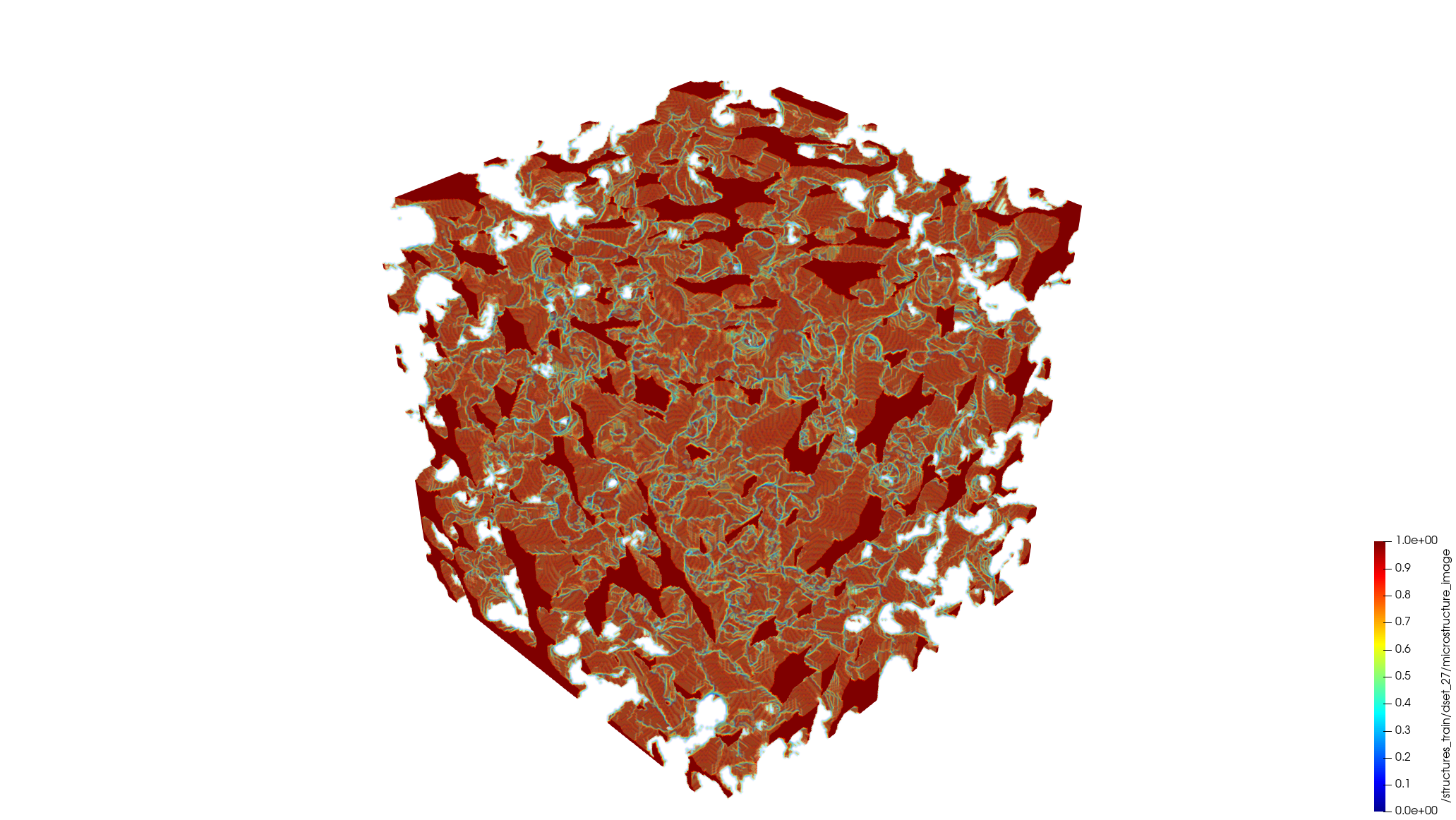}} & 
        \subcaptionbox{Spinodal system                      \label{fig:subfig5}}{\includegraphics[width=0.235\textwidth,trim={17cm 0 17cm 3cm},clip]{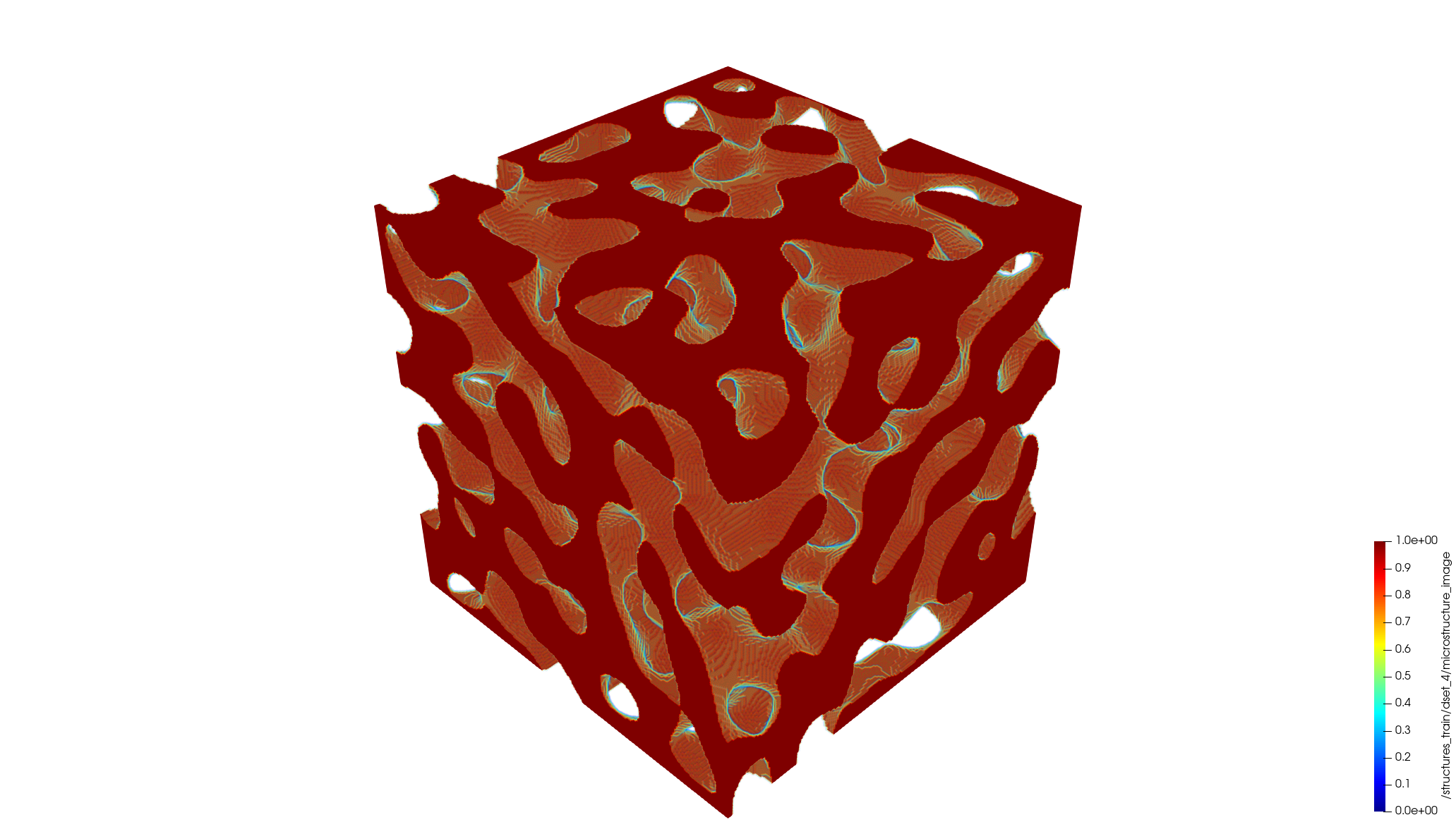}} \\[.7cm]
        \subcaptionbox{Hard ellipsoids                      \label{fig:subfig6}}{\includegraphics[width=0.235\textwidth,trim={17cm 0 17cm 3cm},clip]{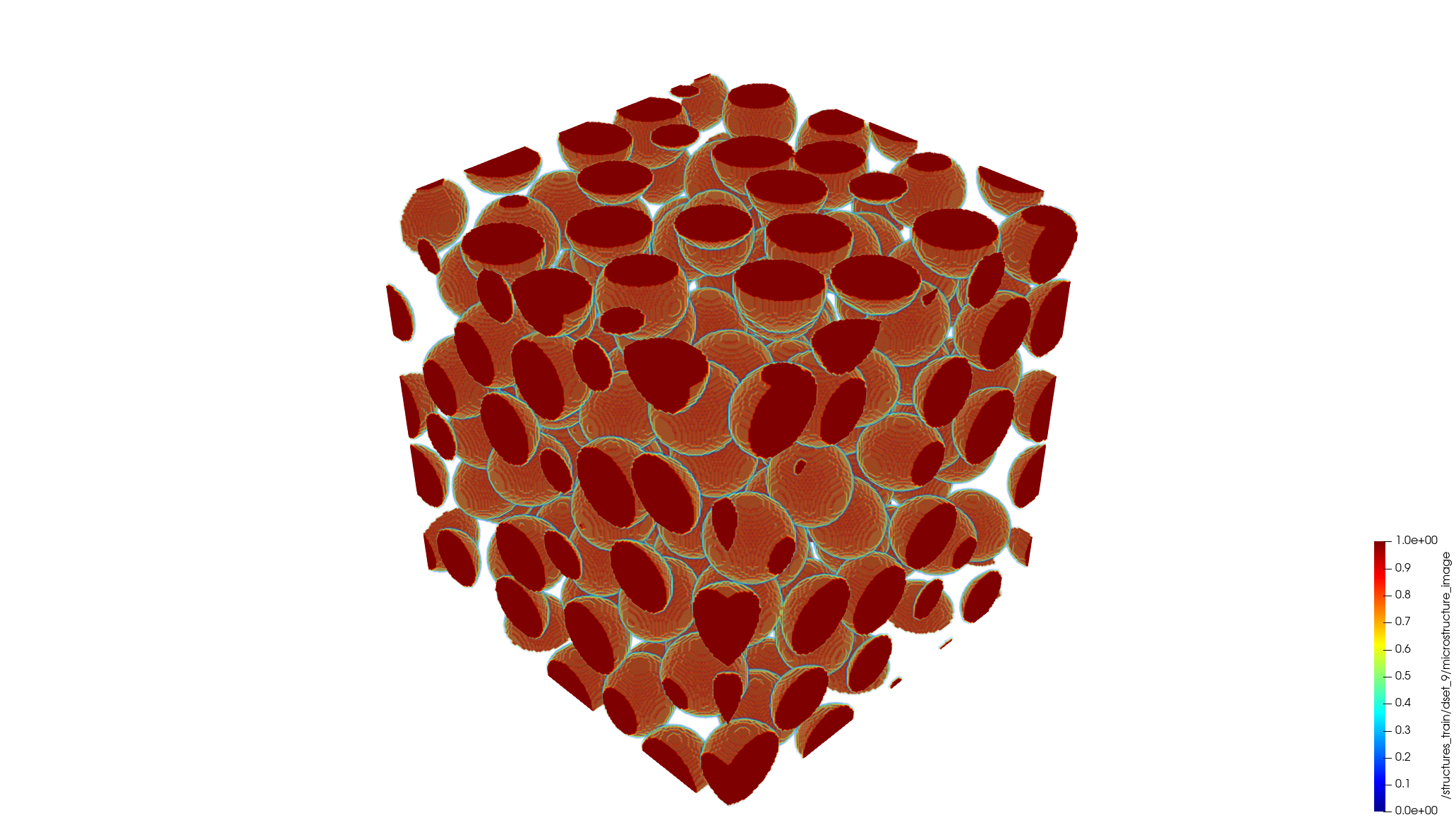}} &
        \subcaptionbox{Smoothed hard ellipsoids             \label{fig:subfig7}}{\includegraphics[width=0.235\textwidth,trim={17cm 0 17cm 3cm},clip]{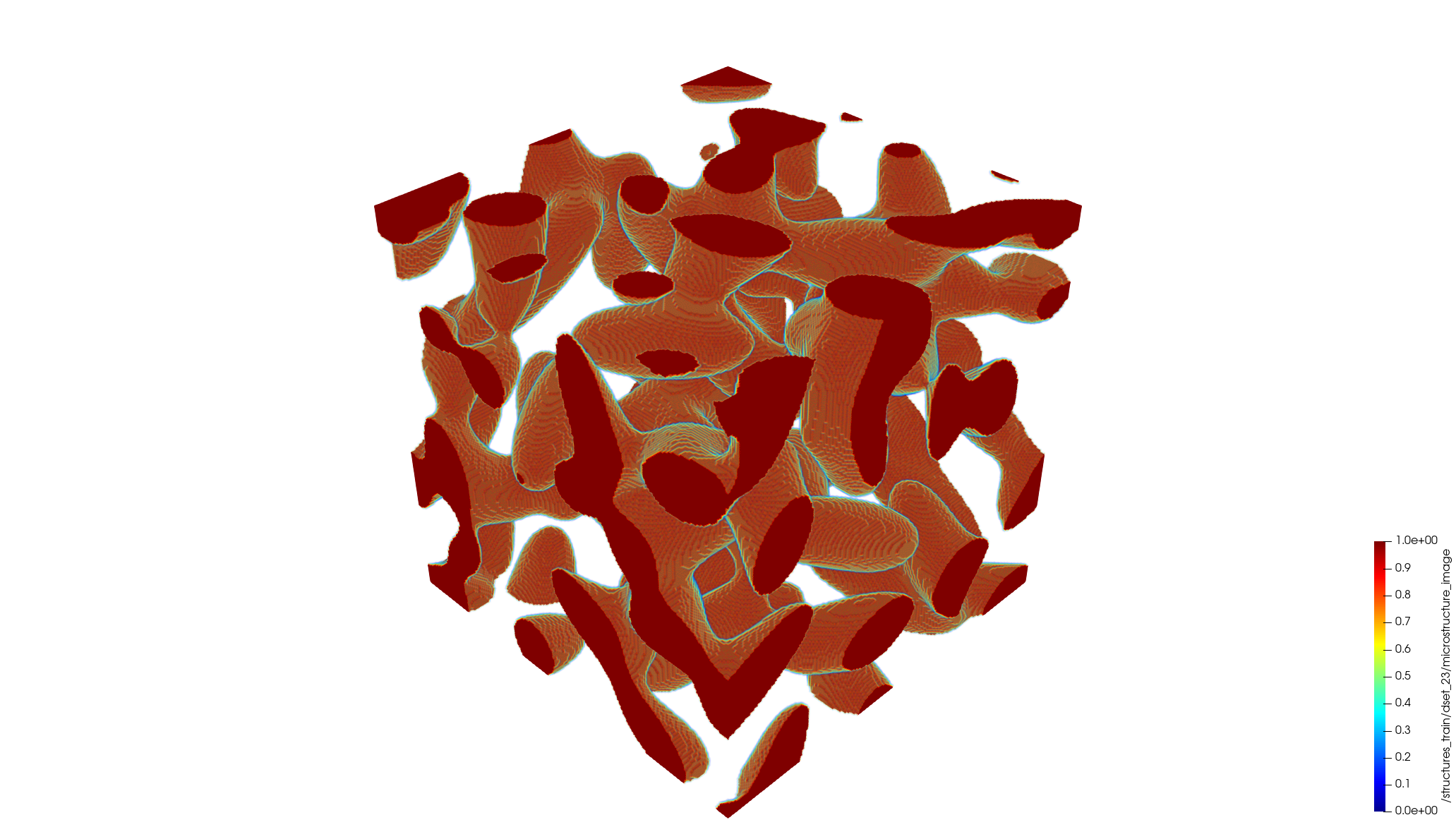}} &
        \subcaptionbox{Soft ellipsoids                      \label{fig:subfig8}}{\includegraphics[width=0.235\textwidth,trim={17cm 0 17cm 3cm},clip]{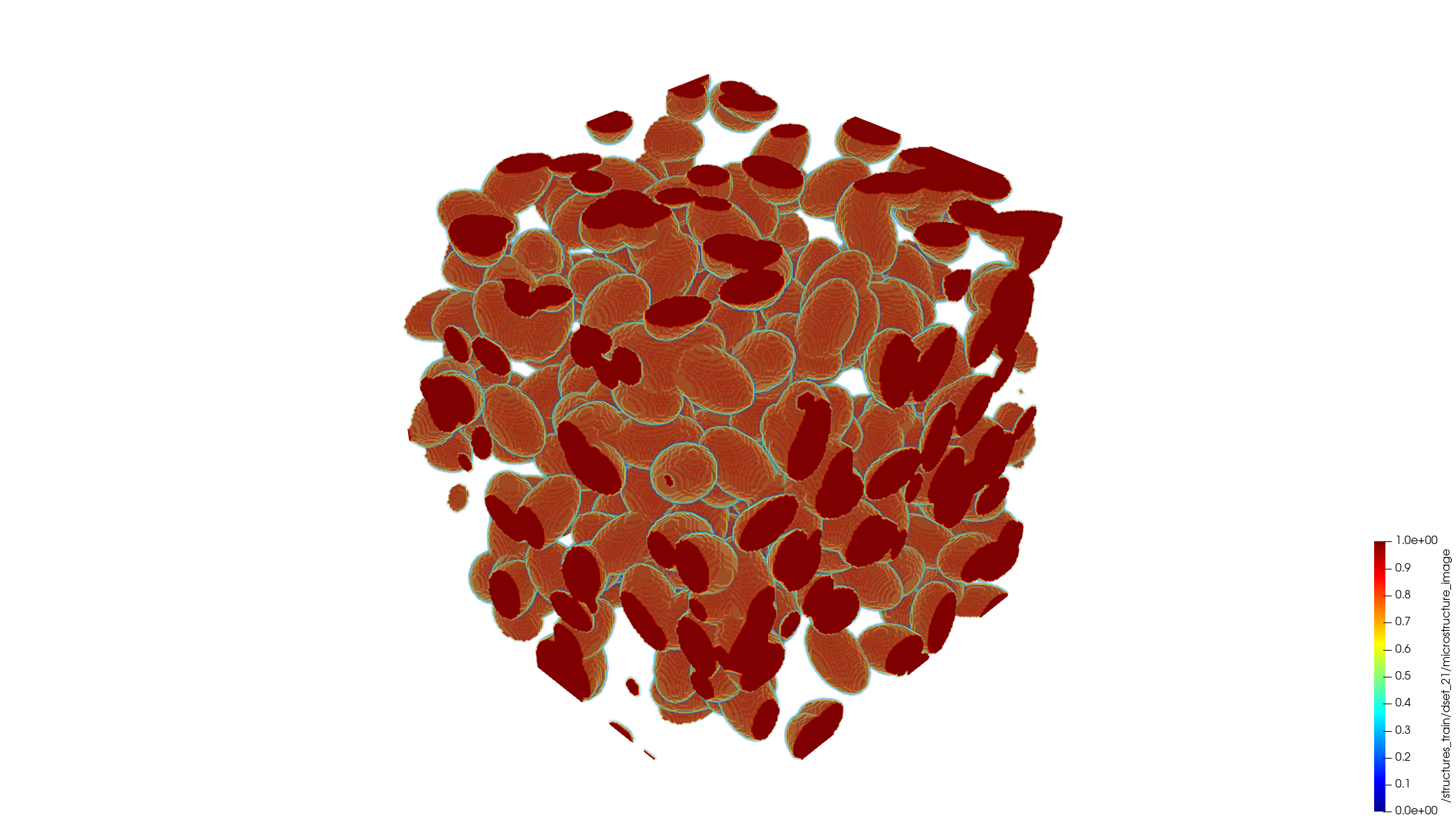}} & 
        \subcaptionbox{Smoothed soft ellipsoids             \label{fig:subfig9}}{\includegraphics[width=0.235\textwidth,trim={17cm 0 17cm 3cm},clip]{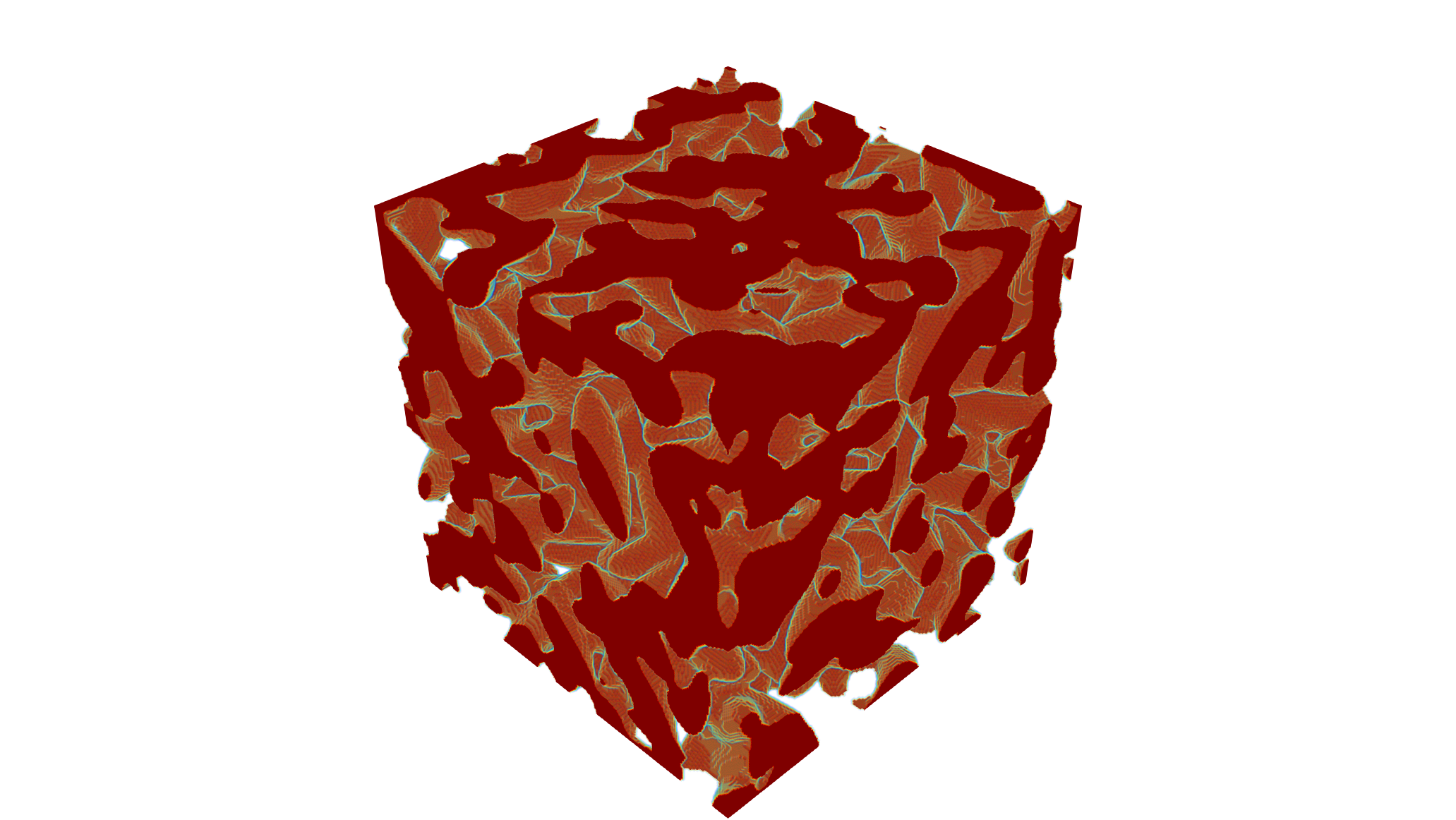}}
    \end{tabular}
    \caption{\protect Examples of 3D microstructures taken from \cite{ulm2020_dataset} for each of the nine classes except for spatial stochastic graphs, with one representative sample shown out of a total of 10,000 samples per class. Note that the inclusion phase is shown in red, and the matrix phase is shown as pore space.}
    \label{fig:example_3D_microstructures}
\end{figure}

In addition to the 2D datasets, we employ a comprehensive collection of synthetic 3D periodic biphasic microstructures published in \cite{ulm2020_dataset}, significantly broadening the scope. These 3D microstructures, each generated at a resolution of $192^3$ voxels, are drawn from nine different classes of stochastic models, ensuring extensive coverage of morphologically distinct structures. Details on the generation process can be found in \cite{Prifling2021}. A total of $90{,}000$ microstructures are considered, partitioned equally among nine classes of,
\begin{enumerate}[label=(\alph*)]
    \item \textbf{Fiber systems:} Randomly placed and oriented fiber inclusions.
    \item \textbf{Channel systems:} Percolating channels forming interconnected pathways.
    \item \textbf{Level sets of Gaussian random fields:} Gaussian random fields thresholded to produce complex matrix-inclusion topologies.
    \item \textbf{Spinodal decompositions:} Structures that mimic phase-separating materials.
    \item \textbf{Hard ellipsoids:} A random sequential adsorption of nonoverlapping ellipsoids.
    \item \textbf{Smoothed hard ellipsoids:} Similar to hard ellipsoids, but with soft boundaries.
    \item \textbf{Soft ellipsoids:} Ellipsoids that may overlap according to the controlled softness parameter.
    \item \textbf{Smoothed soft ellipsoids:} Similar to soft ellipsoids, but with softened boundaries.
    \item \textbf{Spatial stochastic graphs:} Graph-based frameworks that govern connectivity patterns in a 3D domain.
\end{enumerate}
Each of these model classes is designed to systematically explore a wide range of volume fractions, anisotropy levels, connectivity, and inclusion-scale length scales. The resulting microstructures exhibit a wide variety of morphological patterns, as illustrated in~\cref{fig:example_3D_microstructures}. The dataset is divided into a training set of $63{,}000$ microstructures and a validation and test set of $13{,}500$ microstructures each. We consider the full 63,000~microstructures of the training set, but only 2,000 out of the in total 27,000 remaining structures are considered for validation (see also \Cref{tab:phase_contrast_sims3D}.

As with the 2D images, the goal is to avoid feeding raw 3D voxel data directly into the surrogate model, which would be computationally prohibitive in several ways: Both training time and, particularly, the memory requirements would be excessive. Furthermore, the inference could require substantial time due to the large number of inputs. In a recent study, a comparison of engineered versus machine-learned features has been carried out by the authors, see~\cite{lissner2024}. In the current study, each 3D microstructure is characterized by a set of 236 statistical descriptors. These descriptors include simple scalar measures such as volume fraction, specific surface area, and constrictivity; distribution-based descriptors like chord length distribution, spherical contact distribution, and geodesic tortuosity distribution; shape-size metrics such as minimal and maximal pore sizes; and topological features like the two-point correlation function. Details on their rigorous definitions and the algorithms for estimating them in periodic 3D images can be found in~\cite{Prifling2021}. The full dataset of 3D microstructures and their associated descriptors is publicly available as part of an open-access dataset~\cite{ulm2020_dataset}.

These descriptors inherently exploit the stationarity and isotropy of the underlying stochastic models. In~\cite{Prifling2021}, these isotropic descriptors were used to predict isotropic effective quantities. However, this is a limitation for predicting possibly anisotropic tensors, such as the effective conductivity tensor $\ull{\ol{\kappa}}$, which may exhibit directional dependence:
We have computed the deviation of the isotropic projection of all of our simulation results and found a median relative Frobenius error of $\sim$ 2\%. Hence, the isotropic descriptors will not be able to fully capture the anisotropic response of the microstructured material. Independent of this deficiency of the inputs, the proposed spectral normalization strategy from \Cref{sec:spectral:normalization} can be applied as it is agnostic to the input representation; see also \Cref{fig:MLmodel}.
In future work, we plan to extend the microstructure descriptor set to include additional anisotropic features that can better capture the directional dependence of the microstructure, which is expected to further reduce the prediction errors.

\subsubsection{3D Linear thermal homogenization surrogate}
A study similar to the 2D problem has been pursued with the following variations:
\begin{itemize}
    \item The input features are different (236 features from \cite{ulm2020_dataset,Prifling2021} for 3D vs. 51~features provided in \cite{lissner2023_dataset,lissner2019}).
    \item The output is a symmetric matrix of size $3\times3$ instead of $2\times 2$, resulting in 6 targets (3D) and 3 targets (2D).
    \item The number of microstructures is higher, see \Cref{tab:phase_contrast_sims3D} vs. \Cref{tab:phase_contrast_sims2D}.
    \item The network architecture was slightly modified. For the 3D problem, 6 hidden layers with $[512, 256, 128, 64, 32, 16]$ neurons and the same mixed set of activation functions were used, i.e., a layer containing 512 neurons was prepended to the existing architecture.
\end{itemize}

\begin{table}[htbp]
    \centering
    \caption{Summary of the datasets for the 3D thermal homogenization problem.}
    \label{tab:phase_contrast_sims3D}
    \begin{tabular}{lccc}
        \toprule
        \textbf{Dataset} & \textbf{Phase contrast} $R$ & \textbf{Microstructures} & \textbf{Samples} \\[0.1cm]
        \midrule
        Train   & $\left\{\dfrac{1}{100},\;\dfrac{1}{50},\;\dfrac{1}{20},\;\dfrac{1}{10},\;\dfrac{1}{5},\;\dfrac{1}{2},\;2,\;5,\;10,\;20,\;50,\;100\right\}$ 
                & 63,000 
                & 756,000 \\[0.3cm]
        Validation & $\left\{\dfrac{1}{100},\;\dfrac{1}{99},\;\dfrac{1}{98},\ldots,\dfrac{1}{3},\;\dfrac{1}{2},\;2,\;3,\ldots,98,\;99,\;100\right\}$ 
                & 2,000 
                & 396,000 \\[0.3cm]
        \bottomrule \\[-0.1cm]
        \textbf{Total} & & \textbf{65,000} & \textbf{1,152,000} \\
        \bottomrule
    \end{tabular}
\end{table}

\begin{figure}[!h]
\centering
\begin{minipage}{.45\textwidth}
  \centering
  \includegraphics[scale=1.1]{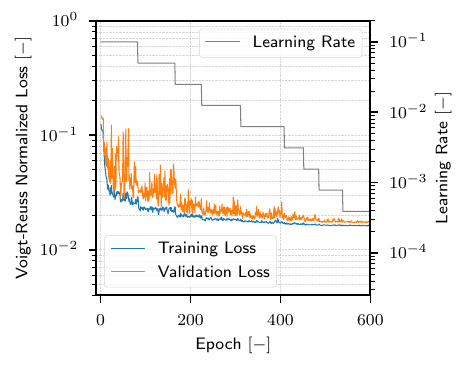}
  \captionof{figure}{Training history of an exemplary training run for the \VRNet{} for the 3D heat conduction problem.}
  \label{fig:therm3d_vrnn_training_history}
\end{minipage}%
\hfill
\begin{minipage}{.45\textwidth}
  \centering
  \includegraphics[scale=1.1]{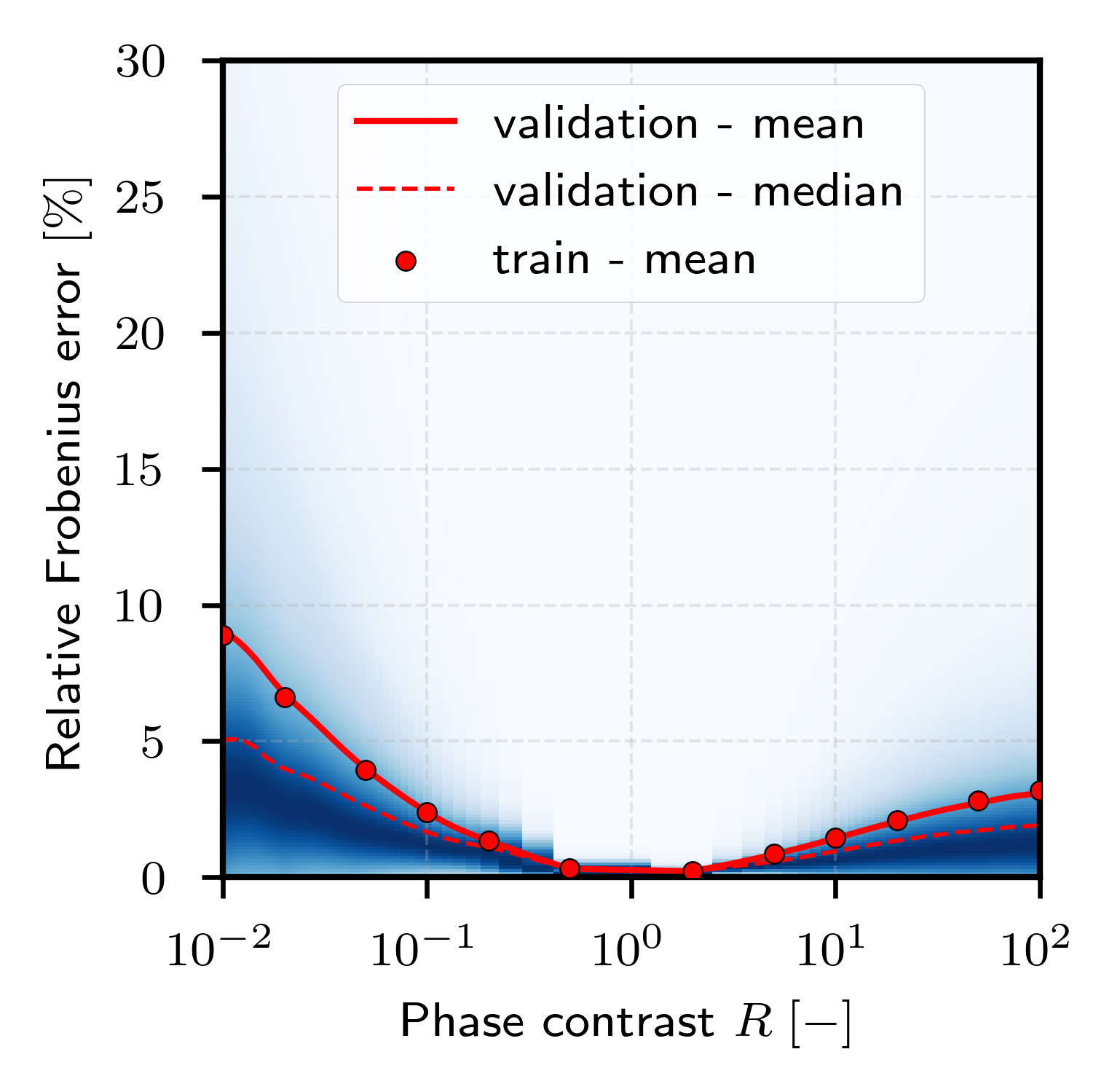}
  \captionof{figure}{Relative error distribution of the Voigt-Reuss net for varying phase contrasts in the 3D heat conduction problem on the validation data.}
  \label{fig:therm3d_rel_error}
\end{minipage}
\end{figure}

The history of loss during an exemplary training run and the relative error distribution for validation and training data are shown in \Cref{fig:therm3d_vrnn_training_history} and \Cref{fig:therm3d_rel_error}, respectively. Compared to the 2D surrogate, the error for $R \gg 1$ appears to be smaller than for $R \ll 1$. This can be explained by less percolation within the microstructures used for 3D than for 2D, i.e., a less pronounced anisotropy and fewer points that are close to either the upper Voigt or the lower Reuss bound.

\begin{figure}[!h]
    \centering
    \includegraphics[scale=1]{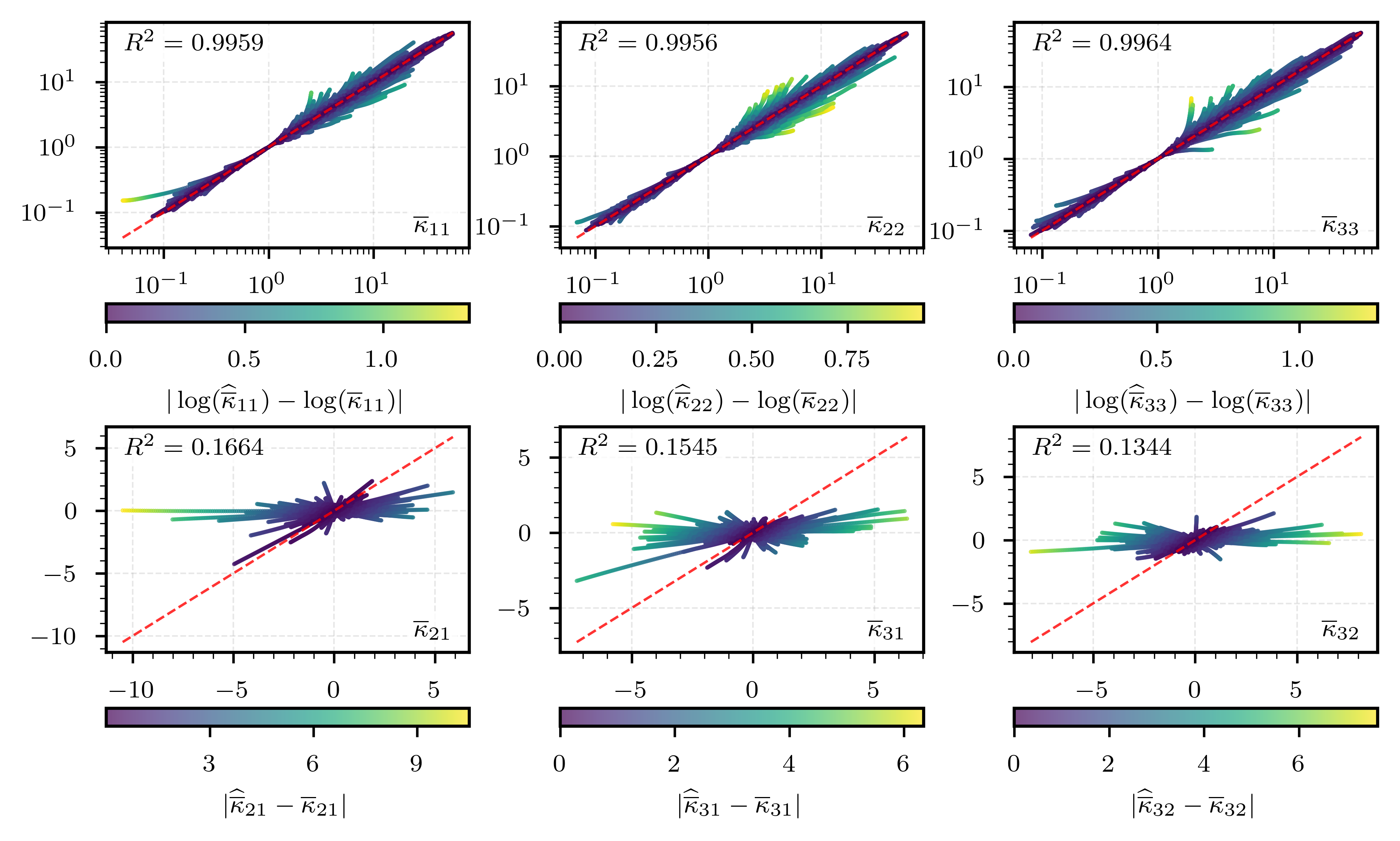}
    \caption{Validation data predictions of the 3D thermal effective conductivity tensor components for the \VRNet{}. The $X$-axis represents the ground truth, and the $Y$-axis shows the predictions.}
    \label{fig:therm3d_predictions_flattened_view}
\end{figure}

\begin{figure}[!h]
    \centering
    \includegraphics[scale=1]{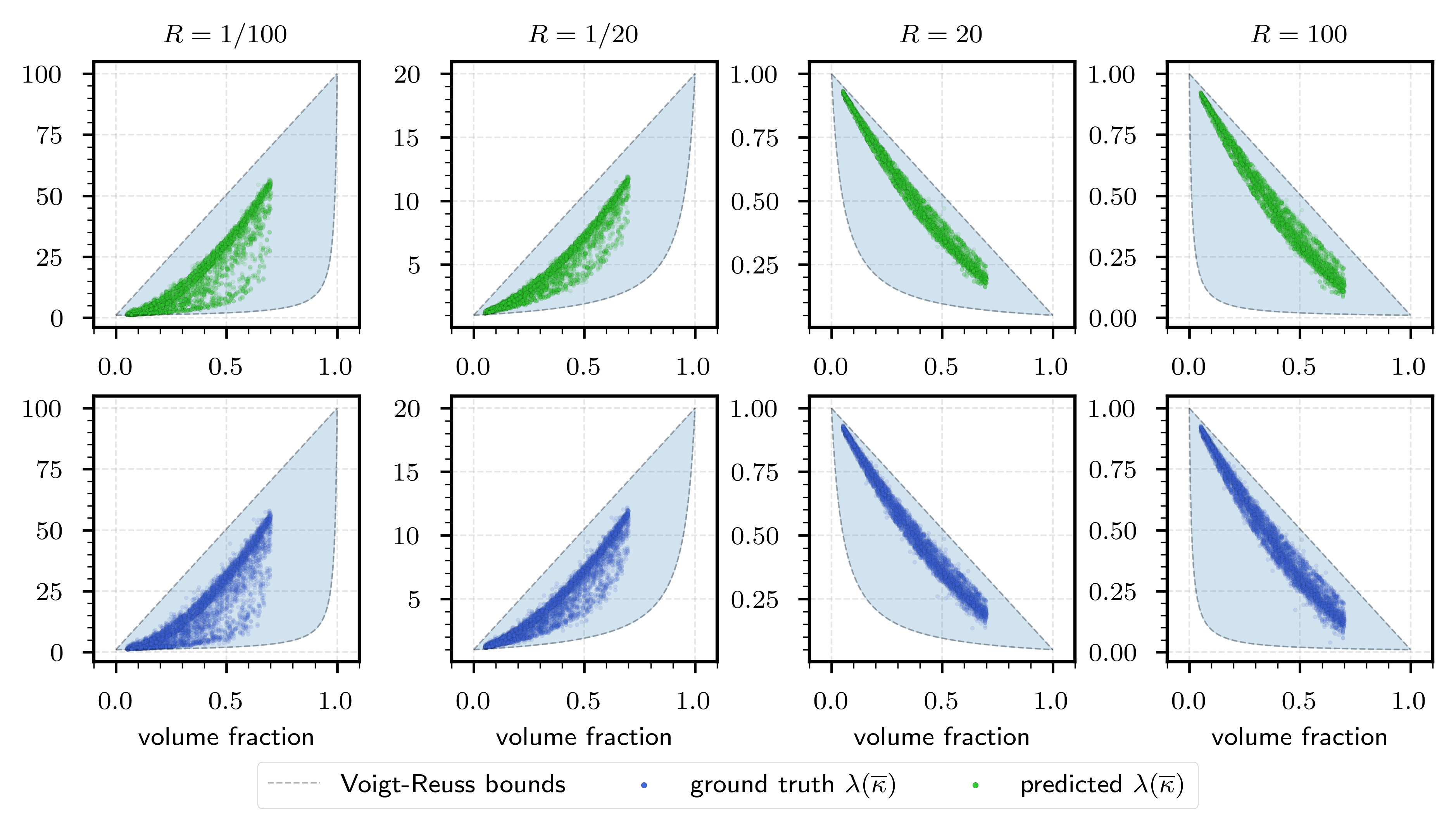}
    \begin{tikzpicture}[overlay, remember picture]
        \node[fill=gray!20, text width=2.75cm, align=center, rotate=90, rounded corners] at (-16.25,7.0) {\VRNet{}};
        \node[fill=gray!20, text width=2.75cm, align=center, rotate=90, rounded corners] at (-16.25,3.25) {Ground truth};
    \end{tikzpicture} 
    \caption{Spectral ground truth and predictions of the effective conductivity tensor for varying phase contrasts for the 3D heat conduction problem.}
    \label{fig:therm3d_predictions_VRbounds}
\end{figure}
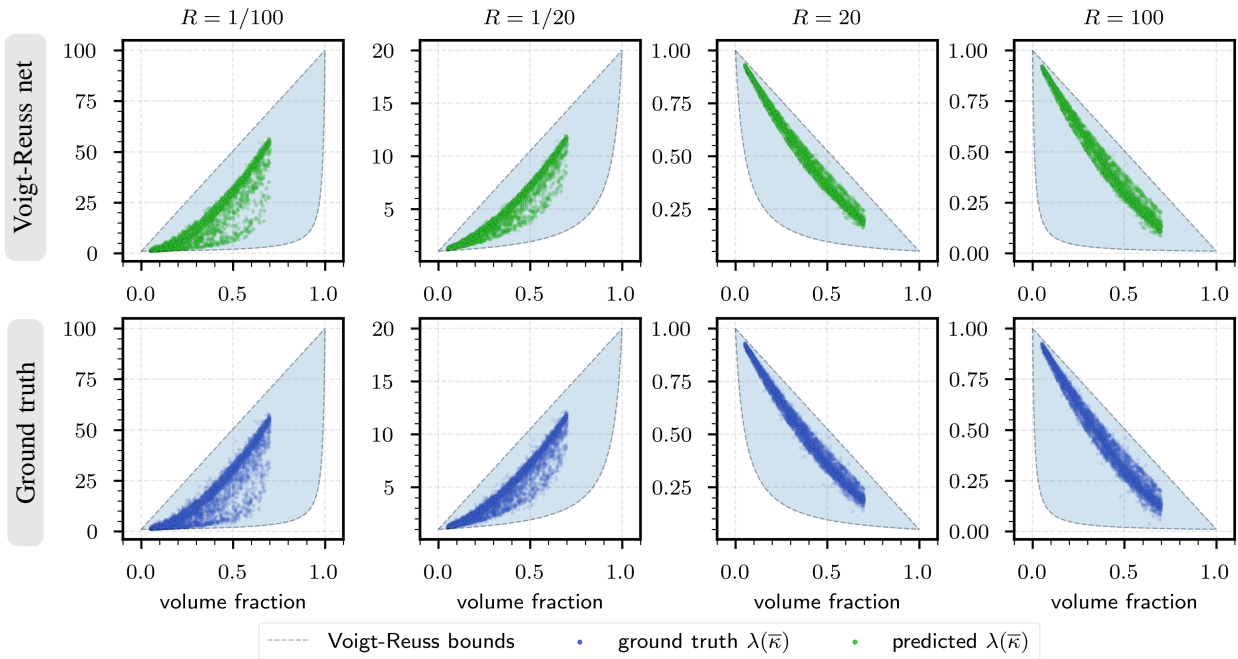

\subsubsection{Performance of the \VRNet{} in data-scarce environments}
In a usual setup, the amount of data used in the current study cannot be matched. For the 3D thermal problems, we solved more than 1 mio. 3D homogenization problems, each comprising three solutions with $192^3\approx7$~mio. degrees of freedom each. Therefore, we investigated the capability of the \VRNet{} to excel in data-scarce environments, i.e., we studied the convergence of the model with respect to the size and type of the training set.

The following scenarios are considered:
\begin{itemize}
\item \textbf{Case O:} The original, i.e., the full training set, is used.
\item \textbf{Case A:} Only four out of the originally 12 different phase contrasts $R\in\{ 10^{-2}, 1/5, 5, 100\}$ were used, and only 20\% of the microstructures were considered at each of these four phase contrasts. This condenses the training set to $6.67$\% of the original size.
\item \textbf{Case B:} We considered all 12 phase contrasts but confined attention to just 1\% of the microstructures, resulting in a dataset that contains only 1\% of the original dataset.
\end{itemize}
Since the effort for training neural networks scales with the dataset size, speed-ups 20 (case \textbf{A}) and 100 (case \textbf{B}) can be achieved by using the pruned datasets, respectively. Similar to the 2D case, a vanilla neural network was trained to directly predict the 6 independent components of the effective conductivity~$\ol{\ull{\kappa}}$ without the use of spectral normalization for all the scenarios (\textbf{O}, \textbf{A}, \textbf{B}).

\begin{figure}[!h]
\centering
\includegraphics[scale=1]{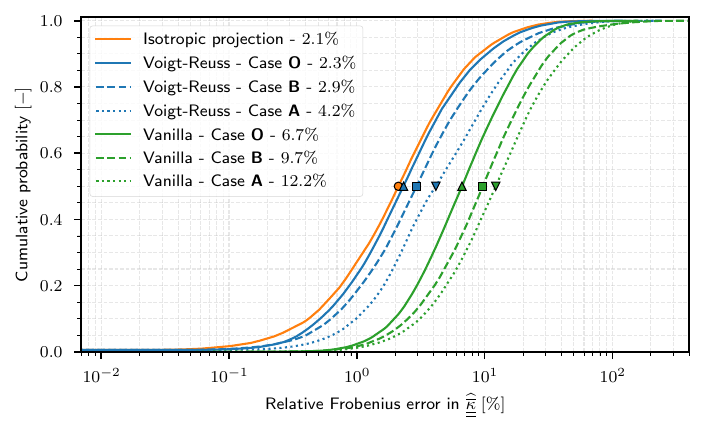}
\caption{Comparison of the empirical cumulative distribution function of the relative Frobenius error of the \VRNet{} (solid line) and the vanilla neural network (dashed line) for the prediction of the 3D thermal validation data (2,000 samples at 198 contrasts). Markers denote the respective median of the relative Frobenius error distribution for each scenario. Three scenarios are compared: original training data (\textbf{O}), $6.67$\% of the total data at 4 phase contrasts $R\in\{ 10^{-2}, 1/5, 5, 100\}$ (\textbf{A}), $1$\% of the total data at all phase contrasts (\textbf{B}).}
\label{fig:therm3d:data:scarce}
\end{figure}

The empirical cumulative distribution function of the relative Frobenius error for scenarios \textbf{O}, \textbf{A}, and \textbf{B} are shown in \Cref{fig:therm3d:data:scarce} for both the \VRNet{} and the vanilla model. For all three scenarios, the error quantiles are consistently factor $\gtapprox 3$ lower. Furthermore, the full dataset used to train the \VRNet{} yields errors that are virtually identical to the distance of the effective conductivity to the space of isotropic conductivities. Since only isotropic features enter as inputs, we consider the \VRNet{} (\textbf{O}) to be quasi-optimal. Interestingly, the use of only 1\% of the microstructures but for all 12 phase contrasts in \textbf{B} outperforms the use of 20\% of the microstructures for just four distinct phase contrasts. This trend is qualitatively and quantitatively the same for \VRNet{} and the vanilla model. One can conclude that the microstructural diversity has less impact than the variation of the phase properties reflected in the contrast~$R$.

\begin{figure}[!h]
    \centering
    \includegraphics[scale=1]{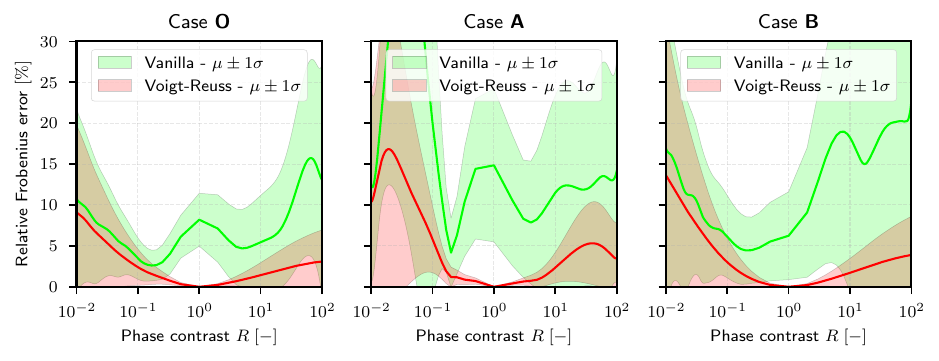}
    \caption{Comparison of Cases (\textbf{O}), (\textbf{A}) and (\textbf{B}) for the \VRNet{} (red) and the Vanilla model (green), solid line shows mean of relative Frobenius error over all validation samples for varying phase contrasts, shaded area shows 1 standard deviation of the relative error distribution for both models. }
    \label{fig:therm3d:data:scarce:model_comparison}
\end{figure}

\section{Closing}

\subsection{Summary}
In this work, we introduce spectral normalization (\Cref{sec:spectral:normalization}) as a powerful tool for strictly enforcing upper and lower bounds on symmetric positive definite operators. The motivation for the spectral normalization stems from the mechanical Voigt \cite{Voigt1889} and Reuss \cite{Reuss1929} bounds, but the same bounds exist (based on the same physical arguments) for homogenized properties that stem from the solution of linear elliptic partial differential equations \citep[e.g.,][]{Wiener1912}. The universal layout of our scheme is shown in \Cref{fig:MLmodel}, where the black box model could be \textit{any} computational procedure, e.g., based on polynomial regression, neural networks, kernel methods, and many more. But the approach is not only agnostic with respect to the black box model but also with respect to the input features: In fact, we merely suggest a clever rescaling requiring a parameterization of orthogonal matrices through parameters~$\ul{\xi}_{\rm q}$.

Based on the general scheme, we construct the \VRNet{}, a neural network-based surrogate employing spectral normalization. In our case, the \VRNet{} was implemented in \texttt{pytorch} using \texttt{torch.nn.parameterizations.orthogonal}. However, concurrent implementations, e.g., in established frameworks such as \texttt{tensorflow}~\cite{Abadi_TensorFlow_Large-scale_machine_2015}, can be realized. On purpose, the \VRNet{} is based on a simplistic network layout: A sequential arrangement of hidden layers with reducing number of neurons is proposed. Concerning the nonlinear activations, we use a slightly non-standard approach that deploys different activations in each layer, including identity mapping, to some of the neurons. No architecture tweaking was done, i.e., no extensive hyperparameter sweeps were performed. This was done on purpose to illustrate the outstanding robustness of the \VRNet{}.

To evaluate the benefits of the \VRNet{} due to spectral normalization, we perform an extensive numerical study. It includes the publication of rich datasets for the community to use in future comparisons and for the construction of benchmarks. We studied hundreds of thousands of 2D and 3D thermal problems simulated with our in-house FFT-based solver FANS \cite{Leuschner2017,combo2022}, which was recently released as open-source software \cite{FANS_github}. The microstructures used in our computations are based on freely available datasets \cite{lissner2023_dataset} (for 2D) and \cite{ulm2020_dataset} (for 3D); see also the related publications \cite{lissner2024, Prifling2021}. The outputs of these simulations are effective thermal conductivity tensors, which obey spectral bounds. They are published in XX\footnote{See Data Availability Statement at the end of the submitted manuscript.}. The phase contrast~$R$, i.e., the ratio of the conductivities of the phases, varied from 10\textsuperscript{-2} to 10\textsuperscript{2}, covering 4 orders of magnitude in property variations.

Input features for the machine learning models were taken from \cite{lissner2024} (2D) in terms of engineered microstructural features and \cite{Prifling2021} (3D); see \Cref{sec:results} for more details. For comparison, we used the same network layout but without spectral normalization to predict the independent components of the conductivity tensor directly. Notably, these components vary massively, i.e., over the order of magnitude, while the \VRNet{} has outputs in the well-confined range $[0, 1]$.

The results for both 2D and 3D datasets confirm that the \VRNet{} has the following properties:
\begin{itemize}
    \item The robustness of the method is excellent, yielding physically meaningful predictions from the first epochs of the training.
    \item The strict enforcement of the bounds is confirmed by the reconstructed data.
    \item Given the same architecture, the orthogonal parameterization and the reconstruction of $\WH{\ullWT{Y}}$ lead to some computational overhead compared to the vanilla neural network for the direct prediction of the components of the effective conductivity.
    \item Even the converged vanilla model supplied with thousands of samples violates the Reuss bound in several situations.
    \item In data-scarce environments, \VRNet{} excels, yielding much better accuracy compared to the vanilla model.
    \item The use of a variety of phase contrasts seems advice given the comparison of the pruned datasets \textbf{A, B} in \Cref{subsubsec:3d_microstructures}: Although the set with pruned $R$-values is much larger in size, it yields worse predictions.
    \item The interpolation in the $R$-range shows excellent accuracy for the \VRNet{} particularly for phase contrasts $R\in[1/10, 10]$. Notably, this is a range relevant for many engineering materials, e.g., for metal matrix composites.
    \item As the conductivity of the "inclusion" phase tends to zero, i.e., $R\to 0$, the \VRNet{} outplays the vanilla network by far, having no outliers.
\end{itemize}

\subsection{Discussion}

The proposed approach is at the confluence of computational homogenization, data-driven surrogate modeling, and physically constrained machine learning. By treating the spectral normalization as a simple data pre-/postprocessing step, the framework remains fully agnostic to the choice of neural network architecture. This flexibility lowers barriers to adoption, facilitating seamless integration into existing digital twins, materials design pipelines, and computational frameworks.

Future studies should explore the very relevant aspect of convergence with respect to dataset sizes more rigorously, trying to find answers to some of the following research questions: What happens for highly anisotropic microstructures? How should the sampling be done most efficiently? Can active learning techniques be applied?

Moreover, the present study was based on thermal conductivity simulations. In the future, the exploration of the mechanical response in the same setup is much awaited. Most importantly, we expect the qualitative and quantitative findings to transfer to this setting. The sampling process, however, will become way more challenging owing to the increased number of material properties to consider (two per phase for isotropic materials), leading to multi-dimensional counterparts of~$R$ from the present study.

While the present study was based on man-made microstructural descriptors, future studies should explore the use of machine-learned features. In that regard, convolutional neural networks and methods from the field of "deep learning" should be explored. In this context, the importance of using small training sets cannot be underrated as they not only reduce compute time but also memory needs during learning, enabling the training on workstations instead of supercomputers, which are required if thousands of microstructures shall be considered in 3D.

On a different note, the use of spectral normalization will be beneficial for many applications in machine learning beyond multiscaling and computational mechanics. For example, any symmetric matrix that has upper and lower bounds in the Löwner sense can be treated with the proposed strategy. The dissemination of the spectral normalization method into the machine learning community, e.g., through the cluster of excellence SimTech, is, therefore, of high relevance. The normalized outputs of the spectral normalization step are also an elegant way of measuring the accuracy of the predictions, which can eliminate unwanted effects due to target quantities covering multiple orders of magnitudes \citep[see also our previous study][]{Fernandez2021}.

\section*{Acknowledgments}
Funded by Deutsche Forschungsgemeinschaft (DFG, German Research Foundation) under Germany's Excellence Strategy - EXC 2075 - 390740016.
Contributions by Felix Fritzen are funded by Deutsche Forschungsgemeinschaft (DFG, German Research Foundation) within the Heisenberg program - DFG-FR2702/10 - 517847245. 
This research was partially funded by the Ministry of Science, Research, and the Arts (MWK) Baden-Württemberg, Germany, within the Artificial Intelligence Software Academy (AISA).
This work is also supported as part of the consortium NFDI-MatWerk, funded by the Deutsche Forschungsgemeinschaft (DFG, German Research Foundation) under the National Research Data Infrastructure - NFDI 38/1 - 460247524. 
We acknowledge the support of the Stuttgart Center for Simulation Science (SimTech).

\subsection*{Financial disclosure}
None reported.

\subsection*{Conflict of interest}
The authors declare no potential conflict of interest.

\section*{Data availability statement}
Upon acceptance of the manuscript, the datasets for the 2D and 3D problem will be published, as well as a simple implementation of the spectral normalization method in \texttt{pytorch}.

\bibliography{literature}

\end{document}